\UseRawInputEncoding

\documentclass[journal]{IEEEtran} 
\newcommand{\abc}[1]{\textcolor{black}{#1}}
\newcommand{\abcn}[1]{\textcolor{black}{#1}}
\newcommand{\abcnn}[1]{\textcolor{black}{#1}} 
\newcommand{\abcnnn}[1]{\textcolor{black}{#1}} 

\newcommand{\Qi}[1]{\textcolor{black}{#1}} 
\newcommand{\NOTEQi}[1]{}

\newcommand{\zq}[1]{\textcolor{black}{#1}} 
\newcommand{\zqq}[1]{\textcolor{black}{#1}} 

\newcommand{\COMMENTS}[1]{}

\newcommand{\rv}{0}
\newcommand{\calF}{{\cal F}}
\newcommand{\calW}{{\cal W}}
\newcommand{\Cat}{\mathrm{Cat}}

\newcommand{\mse}{\mathrm{mse}}
\newcommand{\mean}{\mathrm{mean}}

\newcommand{\up}{\mathrm{up}}

\newcommand{\citet}[1]{\citeauthor{#1}\shortcite{#1}}
\newcommand{\citep}{\cite}

\usepackage{multirow}

\usepackage{times}
\usepackage{epsfig}
\usepackage{graphicx}
\usepackage{amsmath}
\usepackage{amssymb}
\usepackage{color}

\graphicspath{{figures/}}

\ifCLASSINFOpdf
\else
\fi
\begin{document}
%
\title{Single-Frame based Deep View Synchronization for Unsynchronized Multi-Camera Surveillance}
%
%
%

\author{Qi~Zhang,~\IEEEmembership{Student Member,~IEEE,}
        and~Antoni B.~Chan,~\IEEEmembership{Senior Member,~IEEE}
\thanks{Qi Zhang is with Guangdong Laboratory of Artificial Intelligence and Digital Economy (SZ), Shenzhen, Guangdong Province, China.

Antoni B. Chan is with the Department of Computer Science, City University of Hong Kong, Tat Chee 83, Hong Kong SAR, China.

E-mail: qi.zhang.opt@gmail.com, abchan@cityu.edu.hk.} 
\thanks{Manuscript received xxx; revised August xxx.}
}

\markboth{IEEE Transactions xxx, ~Vol.~xx, No.~xx, xx~xx}%
{Zhang \MakeLowercase{\textit{et al.}}: IEEE Transactions xxx}

%



\maketitle


\begin{abstract}
  Multi-camera surveillance has been an active research topic for understanding and modeling scenes. Compared to a single camera, multi-cameras provide larger field-of-view and more object cues, and the related applications are multi-view counting, multi-view tracking, 3D pose estimation or 3D reconstruction, \emph{etc}. It is usually assumed that the cameras are all temporally synchronized when designing models for these multi-camera based tasks. \abcnn{However, this assumption is not always valid, especially for multi-camera systems with network transmission delay and low frame-rates due to limited network bandwidth, resulting in desynchronization of the captured frames across cameras.}
  To handle the issue of unsynchronized multi-cameras, in this paper, we propose a synchronization model that works in conjunction with existing DNN-based multi-view models, thus avoiding the redesign of the whole model.
  We consider two variants of the model, based on where in the pipeline the synchronization occurs, scene-level synchronization and camera-level synchronization. The view synchronization step and the task-specific view fusion and prediction step are unified in the same framework and trained in an end-to-end fashion. Our view synchronization models are applied to different DNNs-based multi-camera vision tasks under the unsynchronized setting, including multi-view counting and 3D pose estimation, and achieve good performance compared to baselines.

\end{abstract}

\section{Introduction}
\par
Compared to single cameras, multi-camera networks allow better understanding and modeling of the 3D world through more dense sampling of information in a 3D scene \cite{aghajan2009multi}. Multi-camera based vision tasks have been a popular research field, especially deep learning related tasks, such as 3D pose estimation from multiple 2D observations  \cite{iskakov2019learnable,Remelli_2020_CVPR}, 3D reconstruction \cite{kar2017learning,huang2018deepmvs}, multi-view tracking \cite{baque2017deep,chavdarova2018wildtrack,Chen_2020_CVPR}, multi-view crowd counting \cite{zhang2019wide}, and ReID \cite{ye2020cross,zhou2019omni,ye2020purifynet,ye2020augmentation,2018Unsupervised}. Usually, it is assumed that the multi-cameras are temporally synchronized when designing DNNs models, \emph{i.e.}, all cameras capture images at the same time point. However, the  synchronization assumption for multi-camera systems may not always be valid \abcnn{in {\em practical applications}} due to a variety of reasons, such as dropped camera frames due to limited network bandwidth or system resources, network transmission delays, \emph{etc}.
Other examples of situations where camera synchronization is not possible include:
\abcnn{1) using images captured from different camera systems;}
 2) using images from social media to reconstruct the crowd at an event; 3) performing 3D reconstruction of a dynamic scene using video from a drone. \abc{Thus, handling unsynchronized multi-cameras is an important issue in the adoption and practical usage of multi-view computer vision.}

\begin{figure}[t]
  \centering
  \includegraphics[width=0.9\linewidth]{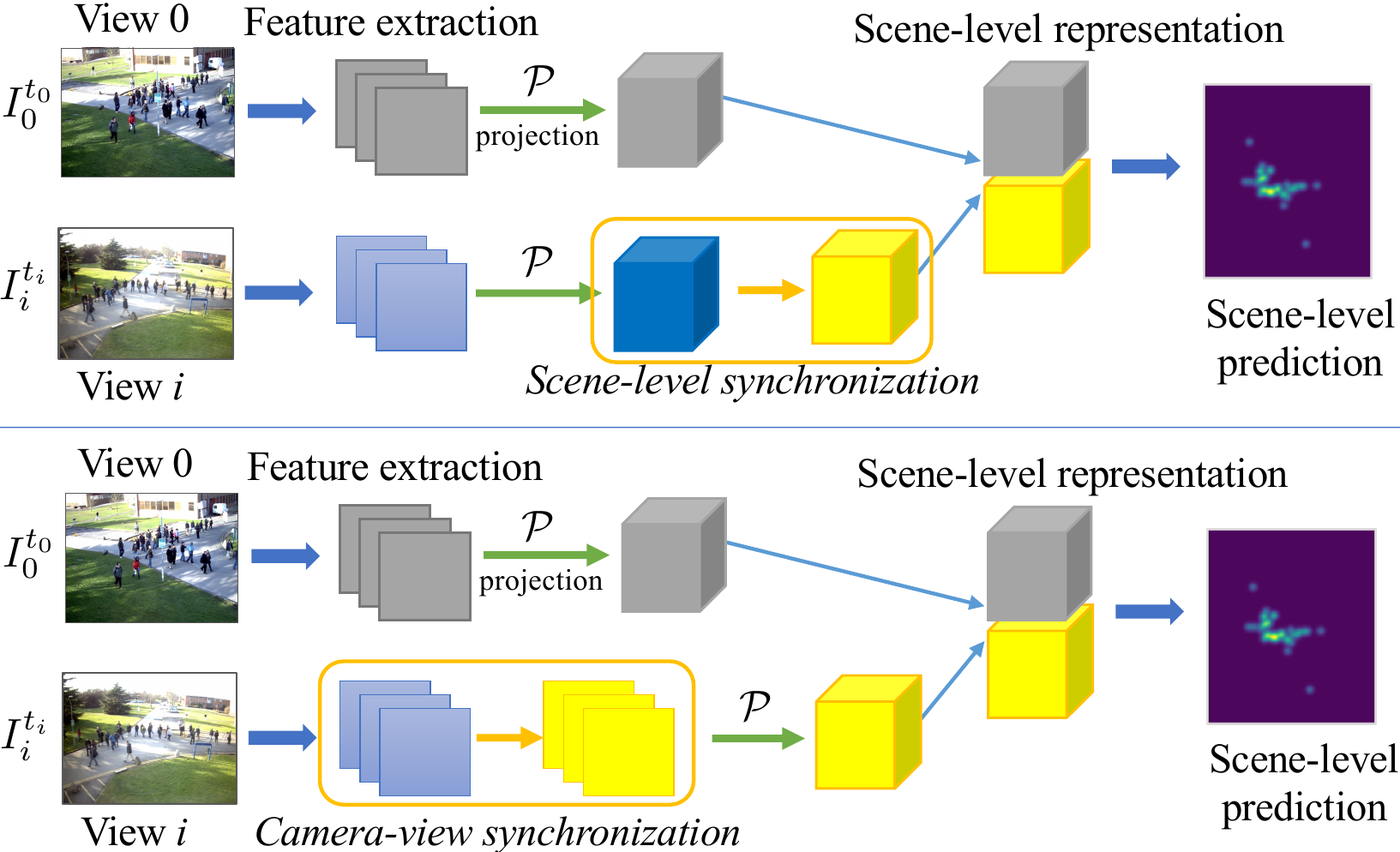}
  \caption{Two variants of the main pipeline for unsynchronized multi-view prediction tasks: \abc{(top) scene-level synchronization is performed after the projection on the scene-level feature representations; (bottom) camera-level synchronization is performed on the camera-view feature maps before projection.}}
\vspace{-0.3cm}
  \label{fig:main_pipeline}
\end{figure}

\begin{figure*}[t]
  \centering
  \includegraphics[width=0.8\linewidth]{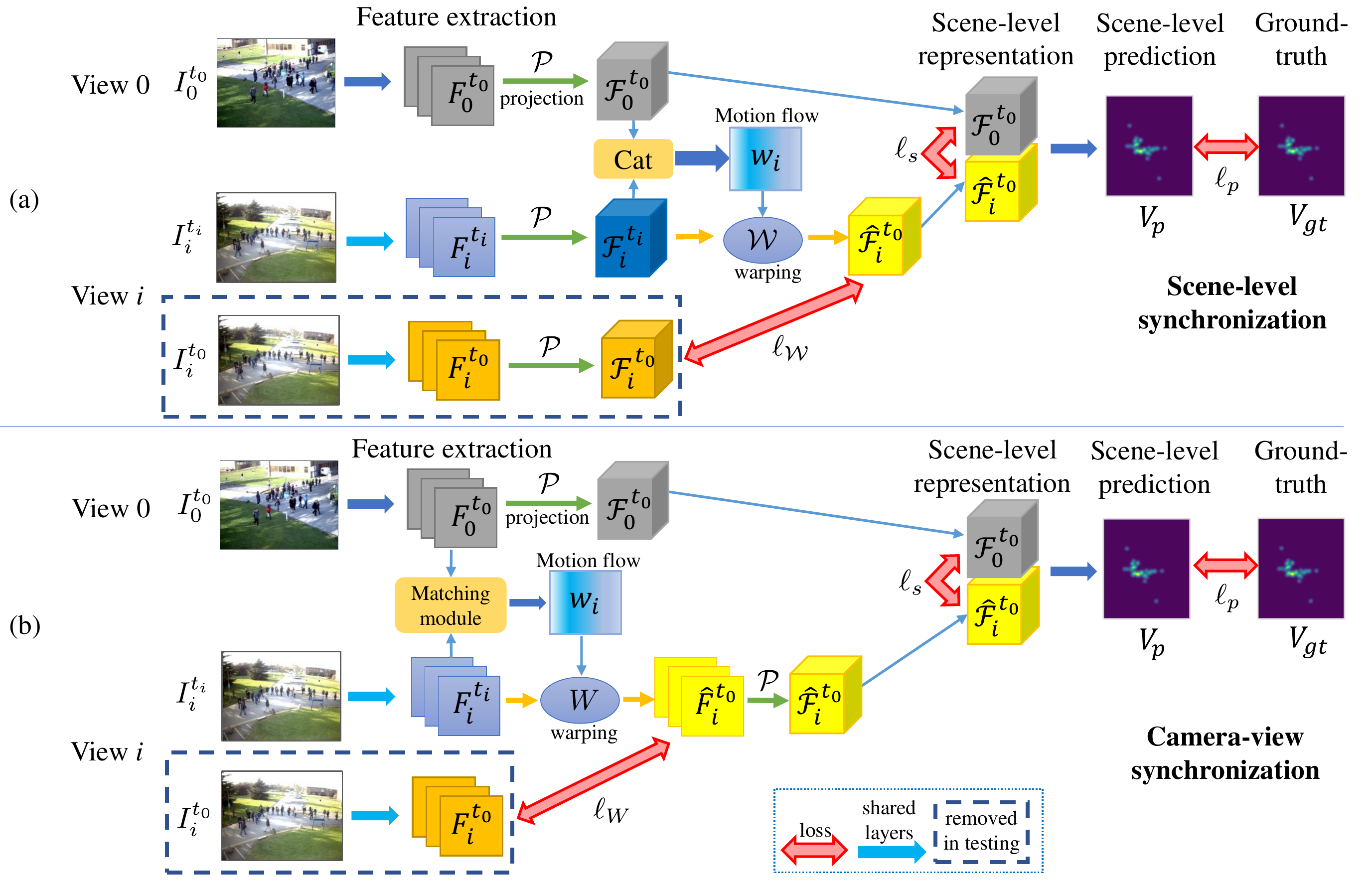}
  \caption{A general multi-view pipeline consists of several stages: \zqq{\emph{camera-view feature extraction, feature projection, multi-view feature fusion to obtain a scene-level representation, and prediction}}.
  (a) {\bf Scene-level synchronization} performs the synchronization \zqq{\emph{after the projection}}. The unsynchronized projected features from the reference view and other views are \zqq{\emph{concatenated to predict the motion flow}}, which is then used to \zqq{\emph{warp the other views' projected features to match those of the reference view}}.
  (b) {\bf Camera-level synchronization} performs the synchronization \zqq{\emph{before the projection}}. The unsynchronized camera-view features from the reference view and other views are \zqq{\emph{matched together
  to predict the motion flow}}, which is used to \zqq{\emph{warp features from other camera views to the reference view}}.
}
\vspace{-0.2cm}
  \label{fig:pipeline}
\end{figure*}

\par
There are several possible methods to fix the problem of unsynchronized cameras. The first method is using hardware-based solutions to synchronize the capture times, such as improving network bandwidth, or by using a central clock to synchronize capture of all cameras in the multi-camera network. However, this will increase the cost and overhead of the system, \abcnnn{and is not possible when there is limited bandwidth}.
The second method is to capture image sequences from each camera, and then synchronize the images afterwards by determining the frame offset between cameras. The fineness of the synchronization depends on the frame rate of the image sequences.
However, this method is not effective when acquiring high frame-rate image sequences is not possible due to limited the bandwidth and storage space, or the frame latency between multi-cameras is random. 
The final method is to modify the multi-view model to handle unsynchronized images, especially for low-frame-rate multi-camera systems or random frame latency between multi-cameras,
such as introducing new assumptions or relaxing the original constraints under the unsynchronized setting.
%
Existing approaches for handling unsynchronized multi-cameras 
are largely based on optimization frameworks \cite{zheng2015sparse,zhang2017dynamics},
but are not directly applicable to DNNs-based multi-view methods, which have seen recent successes in tracking \cite{baque2017deep,chavdarova2018wildtrack}, 3D pose estimation \cite{iskakov2019learnable}, and crowd counting \cite{zhang2019wide,zhang2020_3d}.


\par
In this paper,
%
%
we propose a synchronization model that operates in conjunction with existing DNN-based multi-view models by using single frames from each camera to deal with low-frame-rate unsynchronized multi-camera systems or random frame latency between multi-cameras.
Our proposed model first synchronizes other views to a reference view using a differentiable module,
and then the synchronized multi-views features are fused and decoded to obtain the task-oriented output.
\zq{
As illustrated in Fig.~\ref{fig:main_pipeline}, the synchronization can either occur after the camera-to-scene (2D-to-3D) projection (Fig.~\ref{fig:main_pipeline} top) or before the projection (Fig.~\ref{fig:main_pipeline} bottom). Thus, to fully explore these options,}
we consider two variants of our model that perform synchronization at different stages in the pipeline (see Fig.~\ref{fig:pipeline}): 1) scene-level synchronization performs the synchronization after projecting the camera features to their 3D scene representation; 2) camera-level synchronization performs the synchronization between camera views first, and then projects the synchronized 2D feature maps to their 3D representations. In both cases, motion flow between the cameras' feature maps are estimated and then used to warp the feature maps to align with the reference view (either at the scene-level or the camera-level).
With both variants, the view synchronization and the multi-view fusion are unified in the same framework and trained in an end-to-end fashion. In this way, the original DNN-based multi-view model can be adapted to work in the unsynchronized setting by adding the view synchronization module, thus avoiding the need to design a new model. Furthermore, the synchronization module only relies on content-based image matching and \abcnn{camera geometry}, 
and thus is widely applicable to many DNNs-based multi-view tasks, such as crowd counting, tracking, 3D pose estimation, and 3D reconstruction.

In summary, the contributions of this paper are 3-fold:
\begin{itemize}
  \item We propose an end-to-end trainable framework to handle the issue of unsynchronized multi-camera images in DNNs-based multi-camera vision tasks. To the best of our knowledge, this is the first study on DNNs-based \abcnn{{\em single-frame} synchronization of multi-view cameras.}
  \item We propose two synchronization modules, scene-level synchronization and camera-view level synchronization,
      which are based on 
      image-based content matching \abcnn{that is guided by epipolar geometry.}
      The synchronization modules can be applied to many different DNNs-based multi-view tasks.
  \item We conduct experiments on multi-view counting and 3D pose estimation from unsynchronized images, demonstrating the efficacy of our approach.
\end{itemize}

The remainder of this paper is organized as follows. We review related works in Section \ref{text:relatedwork}. In Section \ref{text:methods}, we propose our single-frame multi-camera synchronization methods, and in Section \ref{text:experiments} we present experiments on two applications, multi-view crowd counting and multi-view 3d human pose estimation.  Finally, Section \ref{text:conc} concludes the paper.

\section{Related Work}
\label{text:relatedwork}

In this section, we review DNN-based methods on synchronized multi-view images and unsynchronized multi-view video tasks, as well as traditional multi-view video synchronization methods. We then review DNN-based image matching and flow estimation methods.


\subsection{DNN-based \abcnnn{synchronized} multi-camera tasks}
Multi-camera surveillance based on DNNs has been an active research area. By utilizing multi-view cues and the strong mapping power of DNNs, many DNNs models have been proposed to solve multi-view surveillance tasks, such as multi-view tracking and detection \cite{2020City,baque2017deep,chavdarova2018wildtrack},  crowd counting \cite{zhang2019wide}, 3D reconstruction \cite{kar2017learning,huang2018deepmvs,choy20163d,Xie_2019_ICCV} and 3D human pose estimation \cite{iskakov2019learnable,kocabas2019self,chen2019unsupervised,joo2015panoptic,pavlakos2017harvesting}. 
\cite{kar2017learning} proposed a deep learning 3D reconstruction framework with differentiable feature projection and unprojection steps.
\cite{ye2020cross} proposed the collaboration ensemble learning for ReID with middle-level sharable two-stream network.
\cite{iskakov2019learnable} proposed volumetric aggregation of feature maps for 3D pose estimation.
The DNN pipelines used for these multi-camera tasks can be generally divided into three stages: the single-view feature extraction stage, the multi-view fusion stage to obtain a scene-level representation, and prediction stage. 
Furthermore, all these DNN-based methods assume that the input multi-views are synchronized, which is not always possible in real multi-camera surveillance systems, or in multi-view data from disparate sources (\emph{e.g.}, crowd sourced images). Therefore, relaxing the synchronization assumption can allow more practical applications of 
multi-camera vision tasks in real world.

\subsection{Tasks on unsynchronized multi-camera \abcnn{video}}

Only a few works have considered computer vision tasks on unsynchronized multi-camera \abcnn{{\em video.}}
\cite{zheng2015sparse} posed the estimation of 3D structure observed by multiple unsynchronized video cameras as the problem of dictionary learning.
\cite{zhang2017dynamics} proposed a multi-camera motion segmentation method using unsynchronized videos by combining shape and dynamical information.
\cite{takahashi2018human} proposed a method of estimating 3D human pose from multi-view videos captured by unsynchronized
and uncalibrated cameras by utilizing the projections of joint as the corresponding points.
\cite{albl2017two} presented a method for simultaneously estimating camera geometry and time shift from video sequences from multiple unsynchronized cameras using minimal correspondence sets.
\cite{kuo2013camera}
addressed the problem of aligning unsynchronized camera views 
by low and/or variable frame rates using the intersections of corresponding object trajectories to match views.

Note that all these methods assume that videos or image sequences are available to perform the synchronization.
In contrast, our framework, \abcnn{which is motivated by practical low-fps systems}, is solving a harder problem, where {\em only a single image} is available from each camera view, \abcnn{\emph{i.e.}, there is no temporal information available}.
Furthermore, these methods pose frame synchronization as optimization problems that are applicable only to the particular multi-view task, and cannot be directly applied to DNN-based multi-view models.
In contrast, we propose a synchronization module that can be broadly applied to many DNN-based multi-camera models, enabling their use with unsynchronized inputs.


\subsection{\Qi{Traditional methods for \abcnn{multi-view video} synchronization}}
\Qi{Traditional synchronization methods usually serve as a preprocessing step for multi-camera surveillance tasks. Except audio-based synchronization like \cite{Hasler2009Markerless}, most traditional camera synchronization methods rely on {\em videos or image sequences} and hand-crafted features for camera alignment/synchronization \cite{Dai2006Subframe,padua2008linear,ChengTri,TresadernVideo,yan2004video}.
%
Typical approaches recover the temporal offset by matching features extracted from the videos, \emph{e.g.}, space-time feature trajectories \cite{CaspiFeature,rao2003view,meyer2008subframe}, image features \cite{imre2012through}, low-level temporal signals based on fundamental matrices \cite{Pundik2010Video}, silhouette motion \cite{Sinha2004Synchronization}, and relative object motion \cite{Gaspar2014Synchronization}.
The accuracy of feature matching is improved using epipolar geometry \cite{imre2012through,Sinha2004Synchronization} and rank constraints \cite{rao2003view}.}
\cite{CaspiFeature} proposed to use the space-time feature trajectories matching instead of feature-points matching to reduce the search space.
\cite{Dai2006Subframe} proposed an iterative procedure to achieve the alignment in space and time with the homography assumption in spatial domain.
\cite{imre2012through} utilized image feature correspondences and epipolar geometry to find the corresponding frame indices and computes the relative frame rate and offset by fitting a 2D line to the index correspondences.
\cite{meyer2008subframe} estimated the frame accurate offset by analysing the trajectories and matching their characteristic time patterns.
\cite{Pundik2010Video} presented a method for online synchronization that relied on the video sequences with known fundamental matrix to compute low level temporal signals for matching.
\cite{rao2003view} proposed the rank constraint of corresponding points in two views to measure the similarity between trajectories to avoid the noise sensitivity of the fundamental matrix.
\cite{Sinha2004Synchronization} proposed a RANSAC-based algorithm that computed the epipolar geometry and synchronization of a pair of cameras from the motion of silhouettes in videos.
\cite{TresadernVideo} estimated possible synchronization parameters via the Hough transform and refined these parameters using non-linear optimization methods.
\cite{yan2004video} relied on correlating space-time interest point distribution in time between videos which represented events in video that had high variation in both space and time.
\cite{Gaspar2014Synchronization} synchronized two independently moving cameras via the relative motion between objects and known camera intrinsic.

%
The main disadvantages for these traditional camera synchronization methods are:
1) Videos and image-sequences are required, which might not be available in practical multi-camera systems with limited network bandwidth and storage;
2) A fixed frame rate of the multi-cameras are usually assumed, which means random frame dropping cannot be handled (except \cite{Pundik2010Video}); 
3) Feature matching is based on hand-crafted features, which lack representation ability, or known image correspondences, which requires extra manual annotations and may not always be available.
Compared with these methods, we consider a more practical and difficult setting: only single-frames and no videos (no temporal information) are available, which means that these traditional video-based methods are not suitable solutions.
These traditional methods perform image content matching using hand-crafted features and traditional matching algorithms, while in contrast our method uses DNN-based image matching. Because we also assume that only single-frames are available, our method also requires DNN-based motion estimation to estimate a frame's features after synchronization.  \abcnn{Finally, our synchronization module is end-to-end trainable with existing multi-view DNNs and thus avoids the redesign of the whole DNNs models to handle unsynchronized multi-cameras.}

\subsection{DNN-based image matching and flow estimation}

Image matching and optical flow estimation both involve estimating image-to-image correspondences, which is related to frame synchronization of multi-views. We mainly review the DNN-based image matching \cite{rocco2017convolutional,phillips2019all,altwaijry2016learning} or optical flow estimation methods \cite{hui2018liteflownet,ilg2017flownet,bai2016exploiting}, which inspire us to solve the unsynchronized multi-camera based problems in a DNN-based fashion.
DNN flow \cite{yu2014dnn} proposed an image matching method based on a DNN feature pyramid in a coarse-to-fine optimization manner. FlowNet \cite{dosovitskiy2015flownet} predicted the optical flow from DNNs with feature concatenation and correlation. SpyNet \cite{ranjan2017optical} combined a classical spatial-pyramid formulation with deep learning and estimated large motions in a coarse-to-fine approach by warping one image to the other at each pyramid level by the current flow estimate and computing an update to the flow.
\cite{rocco2017convolutional} addressed image correspondence problem using a convolutional neural network architecture that mimics classic image matching algorithms. PWC-Net \cite{sun2018pwc} uses a feature pyramid and one image feature map is warped to the other at each scale, which is guided by the upsampled optical flow estimated from the previous scale. \cite{lai2019bridging} proposed a single 
network to jointly learn spatiotemporal correspondence for stereo matching and flow estimation.


\par
Our method is related to the DNN-based image matching and optical flow estimation, but the difference is still significant:
1) Typical image/geometric matching only involves {\em either} a camera view angle transformation (\emph{e.g.}, camera relative pose estimation, \abcn{stereo matching}) {\em or} a small time change in the same view (optical flow estimation), while {\em  both} factors appear in our problem, which makes our problem harder;
2) Image/geometric matching is directly supervised by the correspondence of two images, while the multi-view fusion ground-truth \Qi{in the 3D world}
is used as supervisory signal in our problem;
3) The 2D-to-3D projection causes ambiguity for multi-view feature fusion, which also causes difficulties for view synchronization.

\section{\abcnn{Single-Frame} DNNs Multi-Camera Synchronization}
\label{text:methods}
\par

In this section we propose our \abcnn{single-frame} synchronization model for DNN-based multi-view models.
The temporal offset between cameras can either be constant latency for each camera (the same offset over time), or random latency (random offsets over time). Similar to most multi-view methods \cite{chavdarova2018wildtrack,zhang2020_3d,Xie_2019_ICCV,iskakov2019learnable},
we assume that the cameras are static and the cameras' intrinsic and extrinsic parameters are known.
The main idea of our method is to choose a camera view as the reference view, and then use the view synchronization model to warp the other camera views to be synchronized with the reference view.
The synchronization model should be general enough to handle both constant and random latencies between cameras, in order to work under various conditions causing de-synchronization.


DNNs models for the multi-camera surveillance tasks typically consist of 3 stages (see in Fig. \ref{fig:main_pipeline}): \textbf{Single-view feature extraction}, which extracts single-view features of the input camera views. \textbf{Multi-view feature projection and fusion}, where a fixed differentiable projection layer is first adopted to project the single-view features to the 3D coordinate map and then the projected multi-view features are fused together to form the scene-level representation.
\zq{The projection layer depends the application task, and our framework can generally handle any differentiable projection layer. For example,
for multi-view counting \cite{zhang2019wide}, the projection maps the 2D camera view to the 3D scene plane at the average person height (assuming all camera pixels fall on the same height plane), while for 3D pose estimation \cite{iskakov2019learnable}, the projection copies features along a view-ray in the 3D grid, assuming an unknown height of each camera-view pixel.
}
\textbf{Prediction}, where the decoder  predicts the final result in the 3D coordinate map, such as ground-plane density maps \cite{zhang2019wide} or 3D reconstruction \cite{kar2017learning}.

In Fig.~\ref{fig:pipeline}, we take multi-view crowd counting \cite{zhang2019wide} as an example to show the pipeline of the proposed single-frame based view synchronization model. In the multi-view fusion model, we denote the input multi-view frames as $\{I_i^{t_0}\}_{i=0}^{n-1}$,  where $i$ denotes the camera view id and $n$ is the input camera view number, and superscript $t_0$ indicates that the frames are all captured at the same time point $t_0$, corresponding to the synchronized multi-camera setup.
After being fed into the single-view feature extractor $F$, the extracted features are denoted as
\begin{align}
F_i^{t_0} = F(I_i^{t_0}), i \in \{ 0, 1, ..., n-1\}.
\end{align}
\zq{For multi-view counting \cite{zhang2019wide}, the projection ${\cal P}$ maps the 2D camera view to the 3D scene plane at the average person height.}
After projection layer $\cal P$, the projected multi-view features are
 	\begin{align}
	\calF_i^{t_0} = {\cal P}(F_i^{t_0}), i \in \{ 0, 1, ..., n-1\}.
	\end{align}
We use $U$ to denote the fusion operation (\emph{e.g.}, concatenation, max-pooling) of the projected multi-view features, thus the fused feature is $U(\calF_0^{t_0}, \ldots, \calF_{n-1}^{t_0})$. Finally, the decoder $D$ is applied to obtain the final prediction $V_p$,
  \begin{equation}	
    \begin{aligned}
	V_p &=  D(U(\calF_0^{t_0}, \ldots,\calF_{n-1}^{t_0} )) 
	\\
       &= D(U({\cal P}(F_0^{t_0}), \ldots, {\cal P}(F_{n-1}^{t_0}))).
	\end{aligned}
  \end{equation}
However, when the input multi-cameras frames are not synchronized, denoted as $\{I_i^{t_i}\}_{i=0}^{n-1}$, the capture time for the $i$-th view $t_i \neq t_0$. Thus, we need to synchronize the camera views first by utilizing the view synchronization model.

The view synchronization model can be embedded into one of the first two stages, synchronizing the extracted single-view features $\{F_i^{t_i}\}$ or projected features $\{\calF_i^{t_i}\}$,
without the need to redesign a new architecture.
Thus, we propose two variants of the synchronization model:
1) scene-level synchronization, where the projected features $\{\calF_i^{t_i}\}$ from different camera views are synchronized during multi-camera feature fusion; and
2) camera-level synchronization, where the camera view features $\{F_i^{t_i}\}$ are synchronized before projection and fusion.
%
We present the details of the two synchronization models next.
Note that we first consider the case when both synchronized and unsynchronized multi-view images are available for training (but not available in the testing stage).
We then extend this to the case when only unsynchronized training images are available.
%


\subsection{Scene-level synchronization}
\par
Scene-level synchronization works by synchronizing the multi-camera features after the projection stage in the multi-view pipeline. The pipeline for scene-level synchronization is shown in the Fig.~\ref{fig:pipeline} (a).
%

\subsubsection{Synchronization module}
Without loss in generality, we choose one view (denoted as view $\rv$) as the reference view, and other views are to be synchronized to this reference view.
We first assume that synchronized frame pairs are available in the training stage. The frames are $I_\rv^{t_\rv}$ from reference view $\rv$ captured at reference time $t_\rv$, and $I_i^{t_\rv}$ and $I_i^{t_i}$ from view $i$ ($i \in \{1, 2, .., n-1\}$) taken at times $t_\rv$ and $t_i$. Note that frames $(I_\rv^{t_\rv}, I_i^{t_\rv})$ are synchronized, while $(I_\rv^{t_\rv}, I_i^{t_i})$ are not.
%

\par
The synchronization module consists of the following stages.
 First, camera frame feature maps $(F_\rv^{t_\rv}, F_i^{t_\rv}, F_i^{t_i})$ (both synchronized and unsynchronized frames) are extracted and projected to the 3D world space, 
 resulting in the projected feature maps $(\calF_\rv^{t_\rv}, \calF_i^{t_\rv}, \calF_i^{t_i})$.
 Second, synchronization is 
 performed between the reference view $\rv$ and each other view $i$.
 The projected feature map $\calF_\rv^{t_\rv}$ from the reference view is concatenated with the projected feature map $\calF_i^{t_i}$ from view $i$,
and then fed into a motion flow estimation network ${\cal M}_s$ to predict the scene-level motion flow $w_i$ between view $i$ \abcnn{at time $t_i$} and the reference view \abcnn{at time $t_0$}:
  \begin{equation}	
    \begin{aligned}
	w_i =  {{\cal M}_s}(\Cat(\calF_\rv^{t_\rv}, \calF_i^{t_i})), i \in \{1, ..., n-1\},
	\end{aligned}
  \end{equation}
where $\Cat$ is the concatenation operation.
The $\calF_i^{t_i}$ from view $i$ is then synchronized with the reference view \abcnn{at time $t_0$} using
a warping transformation $\calW$ 
guided by $w_i$, $\calW(w_i, \calF_i^{t_i})$,
  \begin{equation}	
    \begin{aligned}
	\hat{\calF}_i^{t_0} =  \calW(w_i, \calF_i^{t_i}), i \in \{1, ..., n-1\},
	\end{aligned}
  \end{equation}
where $\hat{\calF}_i^{t_0}$ are the warped projected features of the $i$-th view synchronized to time $t_0$.
\zq{
Note that the warping $\cal W$ only applies spatial shifting to the feature map ${\cal F}_i^{t_i}$, i.e., it only changes the feature locations and does not change the feature values.
}
Finally, the reference view features $\calF_\rv^{t_\rv}$ and estimated warped features of the other views
$\{\hat{\calF}_i^{t_0}\}$
are fused and decoded to obtain the final scene-level prediction $V_p$:
  \begin{align}	
	V_p &=  D(U(\calF_0^{t_0},\hat{\calF}_1^{t_0}, \ldots, 
	 \hat{\calF}_{n-1}^{t_0} )) \\
        &=  D(U(\calF_0^{t_0},\calW(w_1, \calF_1^{t_1}), \ldots, {\calW}(w_{n-1}, {\calF}_{n-1}^{t_{n-1}}) )).
  \end{align}
In the testing stage, only unsynchronized frames $(I_\rv^{t_\rv}, I_i^{t_i})$ are available and the forward operations related to frame $I_i^{t_\rv}$ are removed from the network.

\par
\subsubsection{Training loss} Two losses are used in the training stage. The first loss is a task-specific prediction loss $\ell_p$ between the scene-level prediction $V_p$ and the ground-truth $V_{gt}$. For example, for multi-view crowd counting $\ell_p$ is the mean-square error, and $V_p, V_{gt}$ are the predicted and ground-truth scene-level density maps.
%
%
The second loss is on the multi-view feature synchronization in the multi-view fusion stage. Since the synced frame pairs are available during training, the feature warping loss $\ell_{\calW}$ encourages the warped features to be similar to the features of the original synced frame of view $i$,
 \begin{equation}	
 \begin{aligned}
	\ell_{\calW}(w_i, \calF_i^{t_\rv},\calF_i^{t_i})  & = \mse(\calF_i^{t_\rv}, \hat{\calF}_i^{t_0}), \\
                                                      & = \mse(\calF_i^{t_\rv}, \calW(w_i, \calF_i^{t_i})),
	\end{aligned}
 \end{equation}
where $\mse$ is the mean-square error loss.
\zq{Note that the warping $\cal W$ only applies spatial shifting, and thus the minimization of the warping loss $\ell_{\cal W}$ in (8) will be based on the feature alignment via scene-level motion flow $w_i$ and not global feature value changes (e.g., color correction).}
Finally, the training loss combines the task loss and the warping loss summed over all non-reference views,
\begin{align}
  \ell & = \ell_p(V_p, V_{gt}) + \gamma\sum_{i=1}^{n-1} \ell_{\calW}(w_i, \calF_i^{t_\rv}, \calF_i^{t_i}),
  \label{scene_training_loss_all}
\end{align}
where $\gamma$ is a hyperparameter.




\subsection{Camera view-level synchronization}

\par

\zq{
Each image pixels' height in 3D space is unknown, and thus the projection operation of multi-camera DNNs models \cite{zhang2019wide,zhang2020_3d,iskakov2019learnable} will either project each pixel to the same assumed height level \cite{zhang2019wide} (causing distortion when the true pixel height is different), or to multiple height levels \cite{zhang2020_3d}, \cite{iskakov2019learnable} (duplicating features along the view ray). These projection cause the features to stretch along the view ray in the 3D scene, which makes their synchronization more difficult due to their imprecise (stretched) and ambiguous (duplicated) nature.}
Therefore, 
we also consider synchronization between camera view features before the projection.
%
The pipeline for camera-level synchronization is presented in Fig.~\ref{fig:pipeline} (b).

\subsubsection{Synchronization model}
%
The view synchronization model is applied to each view separately. The camera view features $(F_\rv^{t_\rv}, F_i^{t_i})$ from the unsynchronized reference view and view $i$ are first passed through a matching module (see below)
and then fed into the motion flow estimation network ${\cal M}_c$ to predict the camera-view motion flow $w_i$ for view $i$. The warping transformation $W$ guided by $w_i$ then warps the camera-view features $F_i^{t_i}$ from view $i$ to be synchronized with the reference view \abcnn{at time $t_0$},  
  \begin{equation}	
    \begin{aligned}
	\hat{F}_i^{t_0} =  W(w_i, F_i^{t_i}), i \in \{1, ..., n-1\},
	\end{aligned}
  \end{equation}
where $\hat{F}_i^{t_0}$ is the warped camera-view features of view $i$ captured at time $t_i$, which is synchronized to reference view $0$ captured at time $t_0$. Finally, the reference and warped camera views are projected
 	\begin{align}
 	\calF_0^{t_0} = {\cal P}(F_0^{t_0}),
	\hat{\calF}_i^{t_0} = {\cal P}(\hat{F}_i^{t_0}), i \in \{1, ..., n-1\},
	\end{align}
and then fused and decoded to
obtain the scene-level prediction $V_p$:
   \begin{align}	
	V_p &=  D(U(\calF_0^{t_0},\hat{\calF}_1^{t_0}, \ldots, 
	 \hat{\calF}_{n-1}^{t_0} )) \\
       &=  D(U({\cal P}(F_0^{t_0}),{\cal P}(\hat{F}_1^{t_0}), \ldots, {\cal P}(\hat{F}_{n-1}^{t_0}))).
  \end{align}
In the testing stage, only unsynchronized frames $(I_\rv^{t_\rv},I_i^{t_i})$ are available and the forward operations related to frame $I_i^{t_\rv}$ are removed from the network.

\subsubsection{Matching module}
\label{sec:matchmod}
\begin{figure}[t]
  \centering
\includegraphics[width=\linewidth]{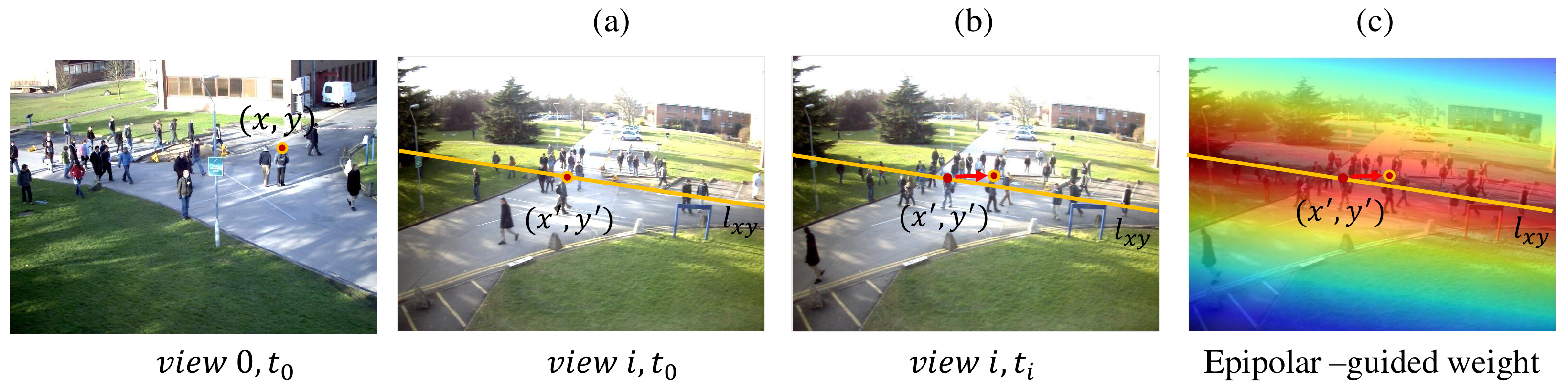}

  \caption{Epipolar-guided weights.
  \abcnn{(a) In the synchronized setting, given the point $(x,y)$ in view $0$, the matched point $(x', y')$ in view $i$ must be on the epipolar line $l_{xy}$. (b) In the unsynchronized setting, we assume a Gaussian motion model of the matched feature location from time $t_0$ to $t_i$. (c) An epipolar-guided weight mask is use to bias the feature matching towards high-probability regions according to the motion model.}
  }


  \vspace{-0.4cm}
  \label{fig:epipolar}
\end{figure}

We propose 3 methods to match features to predict the view-level motion flow.
The first method concatenates the features $(F_\rv^{t_\rv}, F_i^{t_i})$ and then feeds them into the motion flow estimation network ${\cal M}_c$ to predict the motion flow $w_i$:
  \begin{equation}	
    \begin{aligned}
	w_i =  {\cal M}_c(\Cat(F_\rv^{t_\rv}, F_i^{t_i})), i \in \{1, ..., n-1\}.
	\end{aligned}
  \end{equation}
The second method builds a correlation map $C_i$ between features from each pair of spatial locations in $F_\rv^{t_\rv}$ and $F_i^{t_i}$,
	\begin{align}
	C_i( (x,y), (x',y') ) = F_\rv^{t_\rv}(x,y)^T F_i^{t_i}(x',y'),
	\end{align}
which is then fed into the motion flow estimation network ${\cal M}_c$ to predict the motion flow $w_i$:
  \begin{equation}	
    \begin{aligned}
	w_i =  {\cal M}_c(C_i), i \in \{1, ..., n-1\}.
	\end{aligned}
  \end{equation}
\abcnn{The third method incorporates camera geometry information into the correlation map to suppress false matches.
If both cameras are synchronized at $t_0$, then according the multi-view geometry, each spatial location in view $\rv$ must match a location in view $i$ on its corresponding epipolar line (Fig.~\ref{fig:epipolar}a). Thus in the synchronized setting, detected matches that are not on the epipolar line can be rejected as false matches.
For our unsynchronized setting, the matched location in view $i$ remains on the epipolar line only when its corresponding feature/object does not move between times $t_0$ and $t_i$. To handle the case where the feature moves, we assume
that a matched feature in view $i$ moves according to a Gaussian motion model with standard deviation $\sigma$ (Fig.~\ref{fig:epipolar}b).
With the epipolar line and motion model, we then build a weighting mask, with high weights on locations with high probability of containing the matched features, and vice versa. Specifically, we set the mask $M_i( (x,y), (x',y') )=1$ if $(x',y')$ is on the epipolar line induced by $(x,y)$, and 0 otherwise, and then convolve it with a 2D Gaussian with standard deviation $\sigma$ (Fig.~\ref{fig:epipolar}c).
We then apply the weight mask $M_i$ on the correlation map $\tilde{C}_i = M_i \odot C_i$, which will suppress false matches that are not consistent with the scene and motion model. Thus, the motion flow $w_i$ is
  \begin{equation}	
    \begin{aligned}
	w_i =  {\cal M}_c(\tilde{C}_i) = {\cal M}_c(M_i \odot C_i), i \in \{1, ..., n-1\}.
	\end{aligned}
  \end{equation}
}

\COMMENTS{
The third method incorporates an additional epipolar constraint between the reference and non-reference views into the correlation map.
 %
For each spatial location in view $\rv$, a corresponding matched location in view $i$ 
must lie on the epipolar line, assuming synchronized cameras. Thus, we can apply a weight mask $M_i$ on the correlation map $\tilde{C}_i = M_i
\odot C_i$, with high weights along the epipolar line and low weights off of the line, which will reject false matches.
\abcn{Specifically $M_i( (x,y), (x',y') )=1$ if $(x',y')$ is on the epipolar line induced by $(x,y)$, and 0 otherwise.}
However, since the cameras are unsynchronized, the actual epipolar line could be ``shifted'' due to the feature moving. Thus, the epipolar mask is convolved with a 2D Gaussian distribution with large standard deviation $\sigma$, thus expanding the epipolar line to handle movements (see Fig.~\ref{fig:epipolar}).
}

\subsubsection{Multi-scale architecture}
\begin{figure}[t]
  \centering
\includegraphics[width=\linewidth]{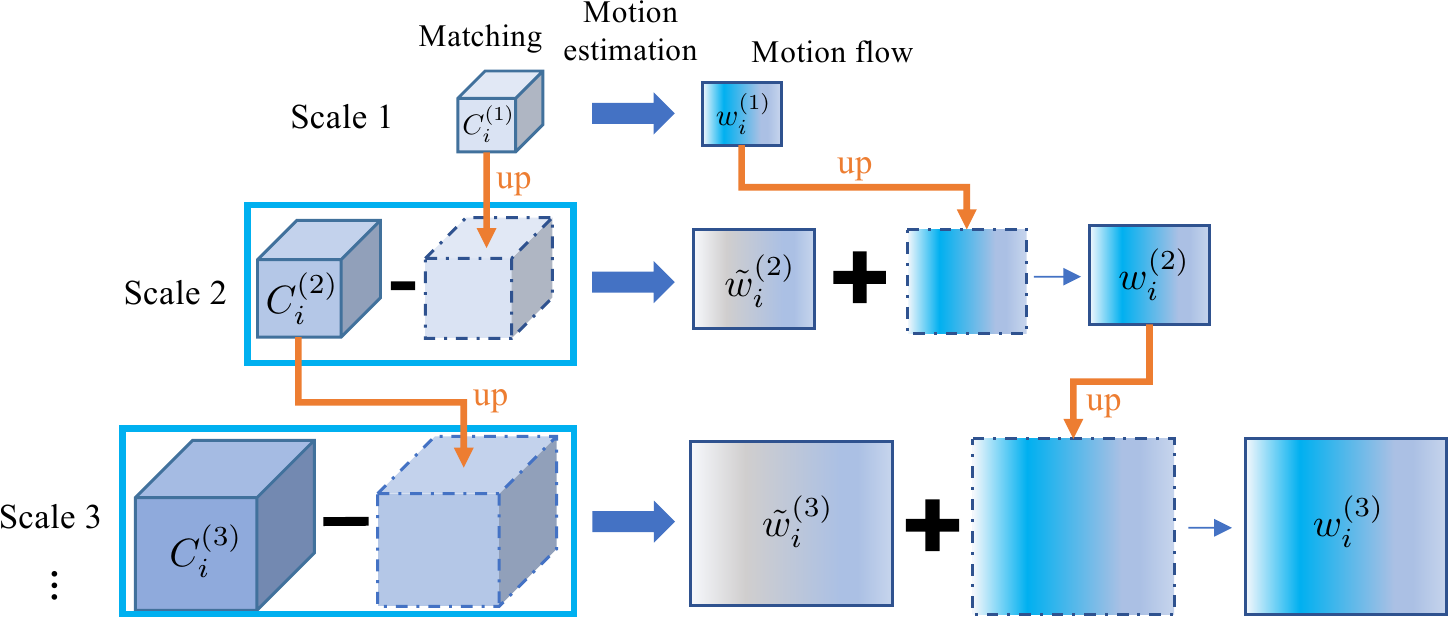}
\caption{\abc{Multi-scale estimation of motion flow.}}
\vspace{-0.5cm}
  \label{fig:multiscale}
\end{figure}

Multi-scale feature extractors are used in multi-camera tasks like crowd counting \cite{zhang2019wide} or to refine the final prediction via multi-scale prediction fusion \cite{lai2019bridging,sun2018pwc}.
Therefore, we next show how to incorporate multi-scale feature extractors with our camera-level synchronization model.\footnote{No extra steps are needed to incorporate multi-scale features with scene-level synchronization because the synchronization occurs after the feature projection.}
 Instead of performing the view synchronization in each scale separately, the motion flow estimate of neighbor scales is fused to refine the current scale's estimate (see Fig.~\ref{fig:multiscale}). In particular, let there be $m$ scales in the multi-scale architecture and $j$ denotes one scale in the scale range $\{1, 2, ..., m\}$, with $m$ the largest scale. The multi-scale predicted motion flow are fused as follows.
\begin{itemize}
  \item When $j=1$ (the smallest scale), the correlation map $C^{(1)}_{i}$ of scale $1$ is fed into the motion flow estimation net to predict the motion flow $w^{(1)}_i$ for scale $1$.
  \item For scales $j>1$, first the difference between the correlation map $C^{(j)}_{i}$ and the upsampled correlation map of the previous scale $\up(C^{(j-1)}_{i})$ is fed into the motion flow estimation net to predict the residual of the motion flow between two scales, denoted as $\tilde{w}^{(j)}_{i}$.
\item The refined motion flow of scale $j$ is 
   \begin{align}
    w^{(j)}_i = \up(w^{(j-1)}_i) + \tilde{w}^{(j)}_i.
   \end{align}
\end{itemize}


\subsubsection{Training loss}
Similar to scene-level synchronization, a combination of two losses (scene-level prediction and feature synchronization) is used in the training stage. The scene-level prediction loss is the same as before.
%
The feature synchronization loss encourages the warped camera-view features at each scale to match the features of the original synchronized frame,
 	\begin{align}
	\ell_{W} & = \mse(F_i^{t_\rv, (j)}, \hat{F}_i^{t_\rv, (j)}) \\
             & = \mse(F_i^{t_\rv, (j)}, W(w_i^{(j)}, F_i^{t_i, (j)})).
	\end{align}
\zq{Similar to scene-level synchronization, the warping function $W$ only applies spatial shifting, and thus the minimization of $\ell_{W}$ in (20) will be based on feature alignment rather than feature value changes.}
Finally, the training loss is the combination of the prediction loss and the synchronization loss summed over all non-reference views and scales,
\begin{align}
  \ell & = \ell_p(V_p, V_{gt}) + \gamma \sum_{i=1}^{n-1} \sum_{j=1}^m \ell_{W}(w_i^{(j)}, \calF_i^{t_\rv, (j)}, \calF_i^{t_i, (j)}),
  \label{camera_training_loss_all}
\end{align}
where $\gamma$ is a hyperparameter.

\subsection{Training with only unsynchronized frames}
In the previous models, we assume that both synchronized and unsynchronized multi-camera frames are available during training. For more practical applications, we also consider the case when only unsynchronized multi-view frames are available for training. In this case, \abc{for the scene-level synchronization}, the warping feature loss $\ell_\calW$ is replaced with a similarity loss $\ell_s$ on the projected features, to indirectly encourage  synchronization of the projected multi-view features,
\begin{align}
  \ell_s = \mean(1 - \cos(\calF_\rv^{t_\rv}, \calW(w_i, \calF_i^{t_i})),
  \label{similarity_loss}
\end{align}
where ``$\cos$'' is the cosine similarity between feature maps (along the channel dimension),
and ``$\mean$'' is the mean over all spatial locations.
\abc{Similarly, for camera-level synchronization, the warping feature loss $\ell_W$ is replaced by the similarity loss of the projected features $\ell_s$.}
\zq{Note that the similarity loss $\ell_s$ is applied after the projection -- thus the warping function only needs to predict the residual motion in the camera view, which is the object motion in time, so as to align the projected features.}



\section{Experiments}
\label{text:experiments}

We validate the effectiveness of the proposed view synchronization model on two unsynchronized multi-view tasks: multi-view crowd counting and multi-view 3d human pose estimation.

\subsection{Implementation details}

\par
The synchronization model consists of two parts: motion estimation network and feature warping layer. The input of the motion estimation network is the unsynchronized multi-view features (the concatenation of the projected features) for scene-level synchronization or the matching result of the 2D camera-view features for camera-level synchronization, and the output is a 2-channel motion flow. The layer setting of the motion estimation network
is shown in Table \ref{table:layer_setting}.
The feature warping layer warps the features from other views to align with the reference views, guided by the estimated motion flow. The feature warping layer is based on the \abc{image resampler} from the Spatial Transformation layer in \cite{jaderberg2015spatial}.

\zq{The synchronized multi-view model consists of feature extraction module, projection module and multi-view prediction module.
For the multi-view counting model \cite{zhang2019wide},  Table \ref{table:FCN-7_fusion_module} shows the model setting of the feature extraction and multi-view prediction module. For the 3D pose estimation model \cite{iskakov2019learnable}, the feature extraction module consists of a ResNet-152 network, a series of transposed convolution layers and a 1 by 1 convolution layer to predict joint heatmaps \cite{2018Simple},
and the V2V-PoseNet \cite{0V2V} is used for multi-view prediction, which is based on hour-glass network \cite{2016Stacked}.
}

\begin{table}[t]
\centering
\caption {The layer settings for the motion estimation net in the view synchronization module. The filter dimensions are output channels, input channels and filter size $w_0\! \times\! h_0\!$.
}
\label{table:layer_setting}
\begin{tabular}{|c|c|}
\hline
Layer & Filter     \\ \hline
conv 1     & $128\!  \times\!  n\!   \times\!  5\!  \times\!  5\! $   \\ 
conv 2     & $128\!  \times\!  128\!  \times\!  5\!  \times\!  5\!$  \\
conv 3     & $64\! \times\!  128\!  \times\!  5\!  \times\!  5\! $ \\
conv 4     & $64\!  \times\!  64\! \times\!  5\!  \times\!  5\! $ \\
conv 5     & $32\!  \times\!  64\!  \times\!  5\!  \times\!  5\! $ \\
conv 6     & $2\!  \times\!  32\!  \times\!  5\!  \times\!  5\! $ \\

\hline
\end{tabular}

\end{table}

\begin{table}[t]
\centering
\caption {The model setting of the synchronized multi-view counting model \cite{zhang2019wide}, consisting of feature extraction and multi-view prediction. 
The Filter dimensions are output channels, input channels, and filter size ($w\!  \times\!  h$).
}
\vspace{-0.2cm}

\label{table:FCN-7_fusion_module}
\begin{tabular}{ll}
\footnotesize
\begin{tabular}{|c|c|}
\hline
\multicolumn{2}{|c|}{Feature extraction} \\ \hline
Layer         & Filter      \\ \hline
conv 1             & $16\! \times\! 1\! \times\!  5\!  \times\!  5$     \\ 
conv 2             & $16\!  \times\!  16\!  \times\!  5\!  \times\!  5$    \\ 
pooling   & $2\!  \times\!  2\!  $         \\ 
conv 3             & $32\!  \times\!  16\!  \times\!  5\!  \times\!  5$   \\ 
conv 4             & $32\!  \times\!  32\!  \times\!  5\!  \times\!  5$ \\ 
pooling   & $2\!  \times\!  2\!  $          \\ 
conv 5             & $64\!  \times\!  32\!  \times\!  5\!  \times\!  5$ \\ 
conv 6             & $32\!  \times\!  64\!  \times\!  5\!  \times\!  5$ \\ 
conv 7             & $1\!  \times\!  32\!  \times\!  5\!  \times\!  5$  \\ \hline
\end{tabular}
&
\footnotesize
\begin{tabular}{|c|c|}
\hline
\multicolumn{2}{|c|}{Prediction}  \\ \hline
Layer & Filter     \\ \hline
   concat   &  - \\ 
conv 1     & $64\!  \times\!  n\!  \times\!  5\!  \times\!  5$   \\ 
conv  2     & $32\!  \times\!  64\!  \times\!  5\!  \times\!  5$  \\
conv 3     & $1\!  \times\!  32\!  \times\!  5\!  \times\!  5$   \\ \hline
\end{tabular}
\end{tabular}
\vspace{-0.2cm}
\end{table}

\subsection{Experiment setup}
\label{text:setup}

We test four versions of our synchronization model: scene-level synchronization (denoted as SLS), and camera-level synchronization using concatenation, correlation, or correlation with epipolar-guided weights (denoted as CLS-cat, CLS-cor, CLS-epi) for the matching module (Section \ref{sec:matchmod}). The synchronization models are trained with the multi-view DNNs introduced in each application later.

We consider two training scenarios: 1) both synchronized and unsynchronized training data is available; 2) only unsynchronized training data is available, which is the more difficult setting.
For the first training scenario, we compare against two comparison methods: {\em BaseS} trains the DNN only on the synchronized data; {\em BaseSU} fine-tunes the BaseS model using the unsynchronized training data (using the full training set). For the second training scenario, {\em BaseU} trains the DNN directly from the unsynchronized data.
\abcnn{Note that traditional synchronization methods \cite{Dai2006Subframe,padua2008linear,ChengTri,TresadernVideo,yan2004video} 
are based on videos (temporal information) and assume high-fps cameras with fixed frame rates, which are unavailable in our problem setting. Thus, traditional and video-based synchronization methods are not suitable for comparison.}

\abc{To test the proposed method, we first create an unsynchonized multi-view dataset from the existing multi-view datasets (the specific datasets are introduced in each application later).}
In particular, suppose the frame sequence in the reference view is captured at times  $t_0 + k{\Delta}t$,
where ${\Delta}t$ is the time offset between neighbor frames, $k \in \{0,\cdots,N-1\}$ and $N$ is the number of frames.
For view $i$, the unsynchronized frames are captured at times $t_0+k{\Delta}t+{\delta}_{i,k}$,
where $\delta_{i,k}$ is the desynchronization time offset between view $i$ and the reference view.
We consider two settings of the desynchronization offset. The first is a {\em constant latency} for each view, ${\delta}_{i,k}=\tau_i$, for some constant value $\tau_i$. The second is {\em random latency}, where the offset for each frame and view is randomly sampled from a uniform distribution, ${\delta}_{i,k} \sim U(-\kappa_i,\kappa_i)$.
%
Finally, since the synchronization is with the reference view, the ground-truth labels for the multi-view task correspond to the times of the reference view, $t_0 + k{\Delta}t$.


\begin{table}[t]
\centering
\caption{Unsynchronized multi-view counting: experiment results for training set with both synchronized and unsynchronized frames. Two desynchronization settings are tested: constant latency and random latency.
\Qi{The evaluation metric is MAE and NAE}.
}
\label{table:result_with_synced}

\begin{tabular}{@{\hspace{0.1cm}}l@{\hspace{0.1cm}}l@{\hspace{0.1cm}}|c@{\hspace{0.1cm}}c@{\hspace{0.1cm}}|c@{\hspace{0.1cm}}c@{\hspace{0.1cm}}|c@{\hspace{0.1cm}}c@{\hspace{0.1cm}}|c@{\hspace{0.1cm}}c@{\hspace{0.1cm}}}

& & \multicolumn{4}{c|}{\em PETS2009} & \multicolumn{4}{c}{\em CityStreet}  \\
& & \multicolumn{2}{c|}{constant} & \multicolumn{2}{c|}{random} & \multicolumn{2}{c|}{constant} & \multicolumn{2}{c}{random} \\
loss & model   & MAE &NAE & MAE &NAE & MAE &NAE & MAE &NAE \\

\hline
\multirow{2}{*}{$\ell_p$}
& BaseS & 7.21&0.200  & 4.58&0.139  &9.07&0.108 & 8.86&0.107
\\
& BaseSU & 4.36&0.137   &4.30&0.140   &9.02&0.106 & 8.82&0.108
\\ \hline
\multirow{4}{*}{$\ell_p,\ell_W$}
& SLS & 4.49&0.145   &4.91&0.154  & 8.23 &0.102  &8.02&0.101
\\
& CLS-cat &4.18& \underline{0.130}   &4.85 &0.150   & 8.82&0.111   &8.57&0.108
\\
& CLS-cor & \underline{4.13}&0.135  &  {\bf 4.03}&\textbf{0.128}  & {\bf 8.03}&\textbf{0.099}  & \underline{7.99}&\underline{0.098}
\\
& CLS-epi & {\bf 3.95}&\textbf{0.130}  & \underline{4.09}&\underline{0.129}  & \underline{8.05}&\underline{0.100} & {\bf 7.93}&\textbf{0.096}
\\ \hline
\end{tabular}

\end{table}

%

\begin{table}[t]
\centering
\caption{Unsynchronized multi-view counting: experiment results for training set with only unsynchronized frames, under constant and random latency.
}
\label{table:result_without_synced}

\begin{tabular}{@{\hspace{0.1cm}}l@{\hspace{0.1cm}}l@{\hspace{0.1cm}}|c@{\hspace{0.1cm}}c@{\hspace{0.1cm}}|c@{\hspace{0.1cm}}c@{\hspace{0.1cm}}|c@{\hspace{0.1cm}}c@{\hspace{0.1cm}}|c@{\hspace{0.1cm}}c@{\hspace{0.1cm}}}

& & \multicolumn{4}{c|}{\em PETS2009} & \multicolumn{4}{c}{\em CityStreet}  \\
& & \multicolumn{2}{c|}{constant} & \multicolumn{2}{c|}{random} & \multicolumn{2}{c|}{constant} & \multicolumn{2}{c}{random} \\
loss & model   & MAE &NAE & MAE &NAE & MAE &NAE & MAE &NAE \\

\hline
$\ell_p$ & BaseU & 6.18 &0.187  &6.22 &0.192   & 10.22&0.134 & 9.35&0.121
\\ \hline
\multirow{4}{*}{$\ell_p,\ell_s$} &
SLS &  5.37&0.178  &4.82&0.150 &  8.50&0.105 & 8.33&0.100
\\
& CLS-cat&  6.00&0.186 &  6.08&0.189 & 8.48&0.102   & 9.17&0.110
\\
& CLS-cor &  {\bf 4.18} &\underline{0.136} & {\bf 4.34}&{\bf 0.136}   & {\bf 8.02}&\underline{0.098} &  \underline{7.77}&\textbf{0.093}
\\
& CLS-epi &   \underline{4.25} &\textbf{0.135}  & \underline{4.77}&\underline{0.144}  &  \underline{8.04}&\textbf{0.095}  &  {\bf 7.70}&\underline{0.094}
\\ \hline
\multirow{4}{*}{$\ell_p$} &
SLS & 7.13&0.226   & 5.30&0.162  & 8.77&0.107 & 8.45&0.107
\\
& CLS-cat & 6.30&0.194  &  5.98&0.184 &   8.28 &\underline{0.098} & 9.15&0.108
\\
& CLS-cor &  {\bf 4.25}&\underline{0.138}   &  {\bf 4.49}&{\bf0.141}    & \underline{8.20}&0.099   &  \underline{8.10}& \underline{0.102}
\\
& CLS-epi &  \underline{4.27}&\textbf{0.135}  &  \underline{4.53}&\underline{0.143}  &{\bf 8.16}&\textbf{0.097}   &  {\bf 7.86} & \textbf{0.096}

\\ \hline
\end{tabular}

\end{table}

\begin{table}[t]
\centering
\caption{Unsynchronized multi-view counting: experiment results for training set with only unsynchronized frames, under constant and random latency and using ground-truth calculated from unsynchronized multi-view frames.
}
\label{table:result_without_synced_unsyncedGT}

\begin{tabular}{@{\hspace{0.1cm}}l@{\hspace{0.1cm}}l@{\hspace{0.1cm}}|c@{\hspace{0.1cm}}c@{\hspace{0.1cm}}|c@{\hspace{0.1cm}}c@{\hspace{0.1cm}}|c@{\hspace{0.1cm}}c@{\hspace{0.1cm}}|c@{\hspace{0.1cm}}c@{\hspace{0.1cm}}}

& & \multicolumn{4}{c|}{\em PETS2009} & \multicolumn{4}{c}{\em CityStreet}  \\
& & \multicolumn{2}{c|}{constant} & \multicolumn{2}{c|}{random} & \multicolumn{2}{c|}{constant} & \multicolumn{2}{c}{random} \\
loss & model   & MAE &NAE & MAE &NAE & MAE &NAE & MAE &NAE \\

\hline
$\ell_p$ & BaseU & 14.89 &0.458  &10.95 &0.484   & 10.96&0.146 & 11.30 &0.149
\\ \hline
\multirow{4}{*}{$\ell_p,\ell_s$} &
SLS &  6.80&0.229  &  6.58 &0.283 &  9.18&0.111 & 9.49&0.117
\\
& CLS-cat&  7.41&0.237 &  6.10&0.237 & 9.72&0.130   & 9.69&0.129
\\
& CLS-cor &  \underline{ 5.91} &\underline{0.201} & \underline{5.93}&\underline{0.240}   & \underline{8.55}&\underline{0.106} &  \underline{8.31}&\underline{0.107}
\\
& CLS-epi &   \bf{5.72} &\textbf{0.184}  & \bf{4.80}&\bf{0.187}  &  \bf{8.32}&\textbf{0.104}  &  {\bf 8.05}&\bf{0.102}
\\ \hline
\multirow{4}{*}{$\ell_p$} &
SLS & 7.85&0.274   & 7.22&0.313  & 9.31&0.109 & 8.91&0.108
\\
& CLS-cat & 7.52&0.240  &  6.20&0.243 &   8.48 &{0.107} & 9.85&0.121
\\
& CLS-cor &  \underline{ 6.98}&\underline{0.244}   &  \underline{6.26}&\underline{0.282}    & \bf{8.03}&\bf{0.099}   &  \underline{8.24}& \underline{0.107}
\\
& CLS-epi &  \bf{6.80}&\textbf{0.229}  &  \bf{5.18}&\bf{0.200}  &\underline{8.23}&\underline{0.102}   &  {\bf 8.16} & \textbf{0.103}

\\ \hline
\end{tabular}

\end{table}

\subsection{Unsynchronized multi-view counting}

We first apply our synchronization model to unsynchronized multi-view counting system, whose bandwidth is assumed to be limited and the frame latency between cameras can be fixed or random.
Here we adopt the multi-view multi-scale fusion model (MVMS) from \cite{zhang2019wide}, which is the state-of-the-art model for multi-view counting DNNs.
We embed the synchronization models in the MVMS model to handle the unsynchronized multi-view frames for crowd counting.

\subsubsection{Datasets and metric}
Two multi-view counting datasets used in \cite{zhang2019wide}, PETS2009 \cite{ferryman2009pets2009} and CityStreet \cite{zhang2019wide}, are selected and desynchronized for the experiments.

{\em PETS2009} contains 3 views (camera 1, 2 and 3), and the first camera view is chosen as the reference view. The input image resolution ($w \! \times \! h$) is  $384\! \times \!288$ and the ground-truth scene-level density map resolution is $152\!\times\!177$. There are 825 multi-view frames for training and 514 frames for testing. The frame rate of PETS2009 is 7 fps ($\Delta t = 1/7 s$). For constant frame latency, $\tau_i \in \{5s, -5s\}$
is used for cameras 2 and 3,
and $\kappa_i=5s$ for random latency.
%

{\em CityStreet} proposed in \cite{zhang2019wide} consists of 3 views (camera 1, 3 and 4), and camera 1 is chosen as the reference view. The input image resolution is $676\!\times\!380$ and the ground-truth density map resolution is $160\!\times\!192$. There are 500 multi-view frames, and the first 300 are used for training and the remaining 200 for testing. The frame rate of CityStreet is 1 fps \Qi{($\Delta t = 1s$)}\footnote{We obtained the higher fps version from the dataset authors.}. 
 %
For constant latency, $\tau_i\in\{3s, -3s\}$ for cameras 3 and 4, 
and $\kappa_i = 3s$ for random latency. 

Following \cite{zhang2019wide}, \Qi{the mean absolute error (MAE) and normalized absolute error (NAE) of the predicted counts on the test set are used as the evaluation metric}:
\begin{align}
  \mathrm{MAE}=\tfrac{1}{N}\sum\nolimits_{i=1}^N|\hat{c}_i-c_i|,
\end{align}
\begin{align}
  \mathrm{NAE}=\tfrac{1}{N}\sum\nolimits_{i=1}^N|\hat{c}_i-c_i|/c_i,
\end{align}
where $c_i$ is the ground truth count and $\hat{c}_i$ is the predicted count, and \zq{$N$ is the number of testing images}.

\subsubsection{Results for training with synced and unsynced frames}
The experiment results using training with synchronized and unsynchronized frames are shown in Table \ref{table:result_with_synced}. The hyperparameter $\gamma = 1$ is used for feature warping loss.
%
\abcn{On both datasets, our camera-level synchronization methods, 
CLS-cor and CLS-epi, perform better than other methods, including the baselines, demonstrating the efficicacy of our approach. Scene-level synchronization (SLS) performs worse than camera-level synchronization methods (CLS), due to the ambiguity of the projected features from multi-views.}
Furthermore after projection to the ground-plane, the crowd movement between frames $I^{t_{\rv}}_i$ and $I^{t_{i}}_i$ on the ground-plane is less salient due to the low resolution of the \abcn{ground-plane feature map}.
CLS-cat performs worse among the CLS methods because simple concatenation of features cannot capture the image correspondence between different views to estimate the motion flow.
%
Finally, the two baselines (BaseS and BaseSU) perform badly on CityStreet because of the larger scene with larger crowd movement between neighboring frames (due to lower frame rate).




\subsubsection{Results for training with only unsynchronized frames} \label{unsynced_frames}
The experiment results by training with only unsynchronized frames (which is a more practical real-world case) are shown in Table \ref{table:result_without_synced}. 
Since the synchronized frames are not available, the MVMS model weights are trained from scratch using only unsynchronized data. Our models are trained with the similarity loss $\ell_s$ (with hyperparameter $\gamma=1000$), which encourages alignment of the projected multi-view features.
Generally, without the synchronized frames in the training stage, the counting error increases for each method. 
Nonetheless, the proposed camera-level synchronization models CLS-cor and CLS-epi 
performs much better than the baseline BaseU.
\abcn{CLS-cor and CLS-epi trained on only unsynchronized data also performs better (on CityStreet) or on par with (on PETS2009) the baseline BaseSU, which uses both synchronized and unsynchronized training data. These two results demonstrate the efficacy of our synchronization model when only unsynchronized training data is available.}
%
%
Finally, the error for almost all synchronization models increases on both datasets when training without the similarity loss ($\ell_p$ in Table \ref{table:result_without_synced}). This demonstrates the effectiveness of using $\ell_s$ to align the multi-view features in training.

\begin{table}[t]
\centering
\caption{Ablation study on the multi-scale architecture of the proposed methods for multi-view counting on CityStreet dataset. The top rows show performance when training with synchronized and unsynchronized frames and using feature warping loss $\ell_W$. The bottom is training only on unsynchronized frames using feature similarity loss $\ell_s$.}
\begin{tabular}{ll|c|c|c|c}
    \hline
 Loss/Training data  & Method   & \multicolumn{2}{c|}{Multi-scale}     & \multicolumn{2}{c}{Single-scale}  \\
    \hline
         &        & MAE  & NAE       & MAE  & NAE \\
    \hline

   &SLS       & 8.02 & 0.101     & 8.31 & 0.100 \\
 $\ell_p,\ell_W$ /
   &CLS-cat   & 8.57 & 0.108     & 8.77 & 0.102 \\
sync and unsync   &CLS-cor   & 7.99 & 0.098     & 8.25 & 0.099 \\
   &CLS-epi   & 7.93 & 0.096     & 8.12 & 0.098 \\
   \hline
   &SLS       & 8.33 & 0.100     & 8.95 & 0.112 \\
$\ell_p,\ell_s$    &CLS-cat   & 9.17 & 0.110     & 9.54 & 0.116 \\
  unsync  &CLS-cor   & 7.77 & 0.093     & 8.62 & 0.111 \\
   &CLS-epi   & 7.70 & 0.094     & 8.59 & 0.110 \\
   \hline

\end{tabular}
\label{tab:scale}
\vspace{-0.2cm}
\end{table}

\subsubsection{Results for using ground-truth from unsynchronized multi-view images}
In the previous experiments (training with only unsynchronized frames, see Sec. \ref{unsynced_frames}), the ground-truth is corresponded (synchronized) to the frames of the reference view. We also perform experiments when the ground-truth scene-level density maps are calculated from the unsynchronized multi-view images. Specifically, we project the same person's image coordinates of each unsynchronized view to the world plane and the average of the projection results is used as the ground-truth person location on the ground. Then, we use the obtained person location map to generate the scene-level density map.

The results for training with ground-truth from unsynchronized multi-view images and only unsynchronized frames can be seen in Table
\ref{table:result_without_synced_unsyncedGT}. From the table, we can also find that the proposed method CLS-cor / CLS-epi can achieve better performance than other methods and CLS-epi achieves the best performance, and the performance can be further improved by adding similarity loss $\ell_s$.

\begin{table}
\centering
\caption{The average feature maps value mean and variance before and after the feature warping of view 2 and 3 of CityStreet dataset.}

\begin{tabular}{l|c|c}
    \hline
    Method   & view 2   & view 3 \\
   \hline
   before warping   & $0.686 \pm 1.006 $      & $0.777 \pm 0.704 $  \\
   after warping     & $0.670 \pm 0.976 $     & $0.761 \pm 0.692 $  \\
   \hline

\end{tabular}
\label{tab:warping-mean-var}
\end{table}

\subsubsection{Ablation study on the multi-scale architecture}
\zq{
%
We next present an ablation study on the multi-scale architecture for the multi-view counting in Table VI.
Generally, the multi-scale architecture performs better than single-scale architecture models, and the proposed CLS-cor/CLS-epi can perform better than SLS or CLS-cat under both single-scale or multi-scale architecture, and under both training paradigms (sync and unsynced, or only unsynced).
}

\begin{table}
\centering
\caption{Comparison of methods on the CityStreet dataset with only unsynchronized frames (both constant and random unsynchronized frames).}

\begin{tabular}{l|c|c}
    \hline
    Method   & constant   & random \\
   \hline
     Color correction     & 8.90/0.108    & 8.64/0.100 \\ 
     \hline
   SLS       & 8.50/0.105        & 8.33/0.100 \\
   CLS-cat   & 8.48/0.102        & 9.17/0.110 \\
   CLS-cor   & \textbf{8.02}/0.098        & 7.77/\textbf{0.093} \\
   CLS-epi   & 8.04/\textbf{0.095}       & \textbf{7.70}/0.094 \\
   \hline

\end{tabular}
\label{tab:color-corr}
\end{table}

\subsubsection{\zqq{Ablation study on color correlation}}
\zqq{
The feature warping module only applies spatial shifting on the features of the unsynced views, i.e., it does not change the values (e.g., color) of the unsynced features (see Eqs. 5 and 10).
To demonstrate this, we calculate the average statistics (mean and variance) of the feature maps before and after feature warping of Views 2 and 3 of CityStreet, and present the results in Table \ref{tab:warping-mean-var}.
The statistics of the feature maps do not change much after performing feature warping, and thus the performance improvement of the feature warping module is not due to color correction (feature value changes).
}

\zqq{
We further perform an ablation study to show that image color correction by itself cannot solve the frame desychronization problem.
On the CityStreet dataset, in the baseline model (MVMS \cite{zhang2019wide}), we add a learnable ``color correction'' layer, comprising an extra 1$\times$1 convolution layer (32 channels) in the branches of the other camera views before the projection and fusion step.
The results are denoted as ``color correction'' in Table \ref{tab:color-corr}.
The error for using ``color correction'' is worse than the proposed SLS, CLS-cor and CLS-epi. The reason is that the desychronization issue comes from the capture time difference between camera views, which is better solved by spatial shifting of features rather than color correction (changing feature values).
}

\subsubsection{Model size and running speed comparison}
\begin{table}
\centering
\caption{The model parameter number and running speed comparison of the baseline methods BaseS/BaseSU/BaseU and the proposed SLC, CLS-cat, CLS-cor and CLS-epi for multi-view counting on CityStreet dataset. The input resolution for the correlation step of the camera-view synchronization module is $160\times95$.}
\begin{tabular}{l|c|c}
    \hline
    Method   & Paras. Num    & FPS \\
   \hline
   BaseS/BaseSU/BaseU    & 853.4K        & 21.9     \\
   \hline
   SLS        & 3.7M   & 8.3     \\
   CLS-cat   & 3.7M        & 8.9 \\
   CLS-cor   & 37.3M        & 7.2 \\
   CLS-epi   & 37.3M        & 3.6 \\
   \hline

\end{tabular}
\vspace{-0.2cm}
\label{tab:speed}
\end{table}

\zq{We present the model size (number of parameters) and running speed of the baseline methods and the proposed SLS, CLS-cat, CLS-cor and CLS-epi in Table \ref{tab:speed}.
The input resolution for the correlation step of the camera-view synchronization module is $160\times95$.
All models are tested on the CityStreet dataset with a Nvidia 1080Ti GPU.
The baseline methods (BaseS, BaseSU and BaseU) do not use view synchronization modules, so their model sizes are smaller and running speeds are faster.
The proposed CLS-cor and CLS-epi methods have the correlation module, and thus have more parameters than SLS or CLS-cat.
CLS-epi is slower than CLS-cor due to the extra multiplication step with the epipolar weights.}

\begin{figure*}[t]
  \centering
\includegraphics[width=0.85\linewidth]{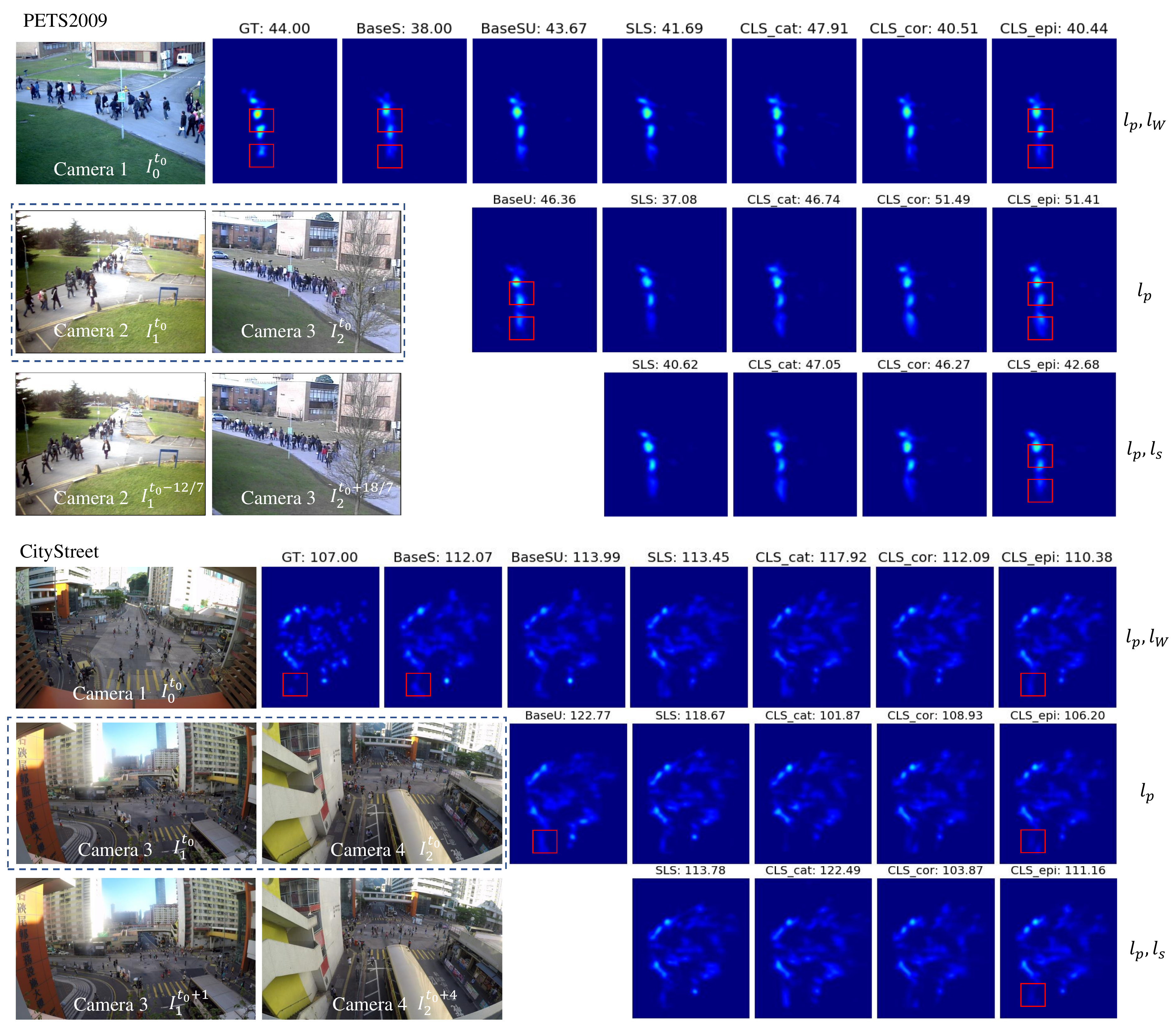}
\caption{\abc{Examples of unsynchronized multi-view crowd counting on PETS2009 (top) and CityStreet (bottom).}
The left shows the input multi-view frames, and note that the synchronized frames (in dotted box) are not used when training with only unsynchronized frames. The input unsynchronized frames are randomly selected around the synchronized frames.
For each dataset, the result of training with synchronized and unsynchronized frames ($l_p$ and $l_W$) is in row 1, the result of training only with unsynchronized frames ($l_p$) is shown in row 2,  and the result of training with unsynchronized frames and using similarity loss between projected features ($l_p$ and $l_s$) is shown in row 3. Generally, 1) the proposed synchronization methods CLS-epi and CLS-cor predict density maps with better quality compared to other comparison methods; 2) the methods achieves better performance when synchronized frames are available in training; 3) when training only with unsynchronized frames, enforcing the similarity loss $l_s$ can help improve the performance.
}
\vspace{-0.4cm}
\label{fig:mv_counting_example}
\end{figure*}

\subsubsection{Visualization results}
Example results are shown in Fig.~\ref{fig:mv_counting_example}. Generally, the proposed synchronization methods CLS-epi and CLS-cor can predict better quality density maps, such as in the red box regions in the figure, where comparison methods tend to over-count these regions due to  the same person being counted multiple times in unsynchronized frames. Furthermore, we also observe that the predicted density map is with better quality when synchronized frames are available compared to training with only unsynchronized frames. Finally, the prediction results are improved if similarity loss is enforced when training with only unsynchronized frames, such as the methods CLS-epi and CLS-cor on PETS2009.

\subsection{Unsynchronized 3D human pose estimation}

We next apply our synchronization model to the unsynchronized 3D human pose estimation task. The DNNs model for  3D human pose estimation  is adopted from \cite{iskakov2019learnable}, which proposed two learnable triangulation methods for multi-view 3D human pose from multiple 2D views: algebraic triangulation and volumetric aggregation. Here we use volumetric aggregation (softmax aggregation) as the multi-view fusion DNN in the experiments.

%
%
%

\begin{table}[t]
\centering
\caption{Unsynchronized 3D human pose estimation: experiment results with random latency. For `CLS-cor' and `CLS-epi', the consistency loss hyperparameter $\gamma=0.01$. \Qi{The evaluation metric is MPJPE and absolute position MPJPE (left/right)}.}
\label{table:results_3d_pose}

\begin{tabular}{l|c|c|c}
\hline
    Latency  & $8/50s$   & $32/50s$ &  $64/50s$  \\
    \hline
      BaseS  & 62.8/59.2 & 78.6/78.2  & 151.1/151.5 \\
      BaseSU & 26.5/27.8  & 49.9/50.1 & 69.4/69.2    \\
      BaseU & 37.3/38.9  & 50.9/50.6 & 71.0/70.7    \\
      CLS-cor($\gamma$=0)  & 25.8/26.9 & \textbf{36.5}/\textbf{36.7} & 56.6/56.9  \\
      CLS-cor              & 25.8/27.0   & 38.2/38.7 & 46.8/47.1                  \\
      CLS-epi         & \textbf{25.7}/\textbf{26.8} & 37.6/37.8 & \textbf{45.7}/\textbf{45.6} \\
\hline
\end{tabular}

\end{table}

\begin{table}[t]
\centering
\caption{Detailed performance for unsynchronized 3D human pose estimation with random latency $\kappa_i = 8/50s$. The evaluation metric is MPJPE.
}

\label{table:results_3d_pose_8}

\begin{tabular}{l@{\hspace{0.1cm}}|c@{\hspace{0.1cm}}c@{\hspace{0.10cm}}c@{\hspace{0.10cm}}|c@{\hspace{0.10cm}}c@{\hspace{0.10cm}}c@{\hspace{0cm}}}
\hline
    Pose              &BaseS  & BaseSU & BaseU  & CLS-cor($\gamma$=0) & CLS-cor & CLS-epi  \\
\hline
    Directions        & 42.8   & 29.3   & 34.3     & 26.1 & 25.8 & 26.1     \\
    Discussion        & 60.7   & 28.4   & 38.8     & 27.3 & 26.7   & 27.0   \\
    Eating            & 60.7   & 26.4   & 28.8     & 23.9  & 24.0 & 23.4  \\
    Greeting          & 63.8 & 19.7    & 32.3      & 25.3 & 24.3 & 25.1 \\
    PhoneCall        & 52.2 & 25.7     & 31.0     & 24.7 & 24.5 & 24.4 \\
    Posing            & 49.7  & 22.0    & 27.6     & 24.1  & 24.0 & 24.0 \\
    Purchases         & 67.5   &24.4    &52.5     & 28.7 & 27.4 & 28.8 \\
    Sitting           & 33.2  & 22.6    & 36.6     & 23.8 &24.0 & 24.0 \\
    SittingDown      & 37.4   & 25.7   & 66.6     & 25.9 & 26.8 & 27.2 \\
    Smoking           & 42.2   & 25.7  & 31.2      & 24.8 & 24.3 & 24.4 \\
    TakingPhoto      & 59.9   & 24.3   & 44.2     & 28.2 & 27.9 & 27.2 \\
    Waiting           & 44.3   & 19.5  & 35.8      & 23.2 & 23.8 & 24.2 \\
    Walking           & 161.1  & 31.9  & 32.1      & 27.0 & 30.2 & 27.8 \\
    WalkingDogs      & 91.5   & 34.2   & 54.8     & 30.1 & 30.1 & 29.8 \\
    WalkingTogether  & 126.8  & 33.9   & 31.8     & 25.5 & 26.8 & 25.5 \\
    \hline
    Average           & 62.8   & 26.5  & 37.3      & 25.8 & 25.8 & 25.7 \\

\hline
\end{tabular}
\vspace{-0.3cm}
\end{table}

\begin{table}[t]
\centering
\caption{Detailed performance for unsynchronized 3D human pose estimation with random latency $\kappa_i = 32/50s$. The evaluation metric is MPJPE.}

\begin{tabular}{l@{\hspace{0.10cm}}|c@{\hspace{0.10cm}}c@{\hspace{0.10cm}}c@{\hspace{0.10cm}}|c@{\hspace{0.10cm}}c@{\hspace{0.10cm}}c@{\hspace{0.10cm}}}
\hline
    Pose     &BaseS  & BaseSU & BaseU &CLS-cor($\gamma$=0) & CLS-cor & CLS-epi  \\
\hline
    Directions        & 46.2  & 48.7   & 65.1 & 42.9 & 42.5   & 43.9     \\
    Discussion        & 75.6  & 53.6  & 55.2 & 38.9 & 41.0   & 41.6   \\
    Eating       & 64.5   & 39.1  & 40.2      & 32.5  & 32.7   & 30.8  \\
    Greeting      & 71.5   & 48.5  & 55.4               & 35.7 & 38.1 & 36.8 \\
    PhoneCall      & 64.5   & 43.6 & 43.0             & 33.9 & 35.2 & 35.1 \\
    Posing       & 49.3   & 42.1 &  43.3         & 32.7 & 33.3 & 30.8 \\
    Purchases     & 111.5  & 50.9 & 48.7           & 35.9 & 42.4 & 40.4 \\
    Sitting       & 55.2   & 46.0 & 43.7        & 33.6 & 33.8 & 34.7 \\
    SittingDown      & 108.3  & 79.3  & 64.9         & 36.8 & 41.8 & 42.8 \\
    Smoking      & 54.5   & 44.3  & 44.0      & 35.5 & 35.9 & 35.7 \\
    TakingPhoto      & 87.9   & 57.0 & 58.6      & 39.3 & 43.0 & 41.3 \\
    Waiting       & 64.3   & 45.6 & 47.3       & 35.5 & 33.7 & 35.0 \\
    Walking       & 150.6  & 47.6 & 48.1         & 34.2 & 37.1 & 34.2 \\
    WalkingDogs     & 123.1  & 66.2 & 67.5          & 44.5 & 49.2 & 49.1 \\
    WalkingTogether     & 125.5  & 50.3  & 52.7     & 36.9 & 38.5 &  34.9 \\
    \hline
    Average    & 78.6   & 49.9  & 50.9      & 36.5 & 38.2 & 37.6 \\

\hline
\end{tabular}
\vspace{-0.3cm}

\label{table:results_3d_pose_32}
\end{table}

\begin{table}[t]
\centering

\caption{Detailed performance for unsynchronized 3D human pose estimation with random latency $\kappa_i = 64/50s$. The evaluation metric is MPJPE.}

\begin{tabular}{l@{\hspace{0.05cm}}|c@{\hspace{0.05cm}}c@{\hspace{0.05cm}}c@{\hspace{0.05cm}}|c@{\hspace{0.05cm}}c@{\hspace{0.05cm}}c@{\hspace{0cm}}}
\hline
    Pose     &BaseS  & BaseSU   & BaseU & CLS-cor($\gamma$=0) & CLS-cor & CLS-epi  \\
\hline
    Directions        & 99.2  & 83.2  & 76.3       & 70.3 & 64.8   & 66.5     \\
    Discussion        & 144.1  & 72.0  & 67.5              & 57.3 & 48.2   & 48.4   \\
    Eating       & 138.2   & 55.3  & 63.2             & 44.7 & 40.4   & 37.9  \\
    Greeting      & 181.3   & 68.0 & 74.1               & 54.8 & 46.9 & 46.3 \\
    PhoneCall      & 138.1   & 58.8 & 61.1                  & 49.2 & 40.7 & 40.5 \\
    Posing       & 121.7   & 53.5 & 50.2                 & 42.1 & 36.6 & 36.3 \\
    Purchases     & 155.7  & 69.0  & 62.4           & 58.1 & 47.0 & 50.6 \\
    Sitting        & 74.0   & 64.2 & 67.8                & 55.2 & 41.1 & 39.6 \\
    SittingDown      & 103.8  & 112.3 & 140.7             & 89.8 & 54.6 & 50.6 \\
    Smoking      & 112.7  & 58.7  & 60.3               & 49.2 & 41.7 & 41.7 \\
    TakingPhoto      & 166.8   & 76.8 & 79.6           & 64.5 & 57.7 & 53.5 \\
    Waiting       & 120.2  & 62.7  & 61.1            & 51.1 & 42.0 & 42.6 \\
    Walking       & 301.1  & 66.3  & 69.2             & 49.7 & 44.0 & 41.7 \\
    WalkingDogs     & 219.2  & 95.9 & 91.6           & 77.0 & 62.7 & 55.5 \\
    WalkingTogether     & 302.7  & 67.2 & 68.9             & 54.5 & 43.5 & 42.5 \\
    \hline
    Average    & 151.1   & 69.4  & 71.0         & 56.6 & 46.8 & 45.7 \\

\hline

\end{tabular}

\label{table:results_3d_pose_64}
\end{table}

\begin{table}[t]
\centering
\caption{Unsynchronized 3D human pose estimation: CLS-epi experiment results with different hyperparameter $\gamma$. The evaluation metric is MPJPE.}

\begin{tabular}
{l@{\hspace{0.30cm}}c@{\hspace{0.30cm}}c@{\hspace{0.30cm}}c}
\hline
    $\gamma$     &0.005  & 0.01 & 0.02  \\
\hline
    $\kappa_i = 8/50s$        & 25.6   & 25.7  & 26.0      \\
    $\kappa_i = 32/50s$       & 38.3   & 37.6  & 37.9    \\
    $\kappa_i = 64/50s$       & 51.7   & 45.7 & 46.8    \\
\hline
\end{tabular}

\label{table:gamma}
\end{table}

\subsubsection{Datasets and Metrics}
We use the Human3.6M \cite{ionescu2013human3} dataset, which consists of 3.6 million frames from 4 synchronized 50 Hz digital cameras along with the 3D pose annotations. We follow the preprocessing step\footnote{https://github.com/anibali/h36m-fetch. Accessed: Oct. 10, 2019.} recommended in \cite{ionescu2013human3}, and sample one of every 64 frames ($\Delta t = 64/50$) for the testing set, and sample one of every 4 frames ($\Delta t = 4/50$) as the training set. The first camera view is always used as the reference view (if the first camera view is missing, the second one is used).
We test desynchronization via random frame latency, with $\kappa_i \in \{8/50, 32/50, 64/50\}$  seconds, \zq{and only use unsynchronized data for training}.
%
%
Following \cite{iskakov2019learnable}, Mean Per Point Position Error (MPJPE) \Qi{and absolute position MPJPE} are used as the metric for evaluation.
%
In training, the single-view backbone uses the pretrained weights from the original 3D pose estimation model. Baseline methods \abcn{BaseS, \zq{BaseSU} and BaseU}
are compared with our proposed camera-view synchronization models CLS-cor and CLS-epi.


%
%

\begin{figure*}[t]
  \centering
\includegraphics[width=0.7\linewidth]{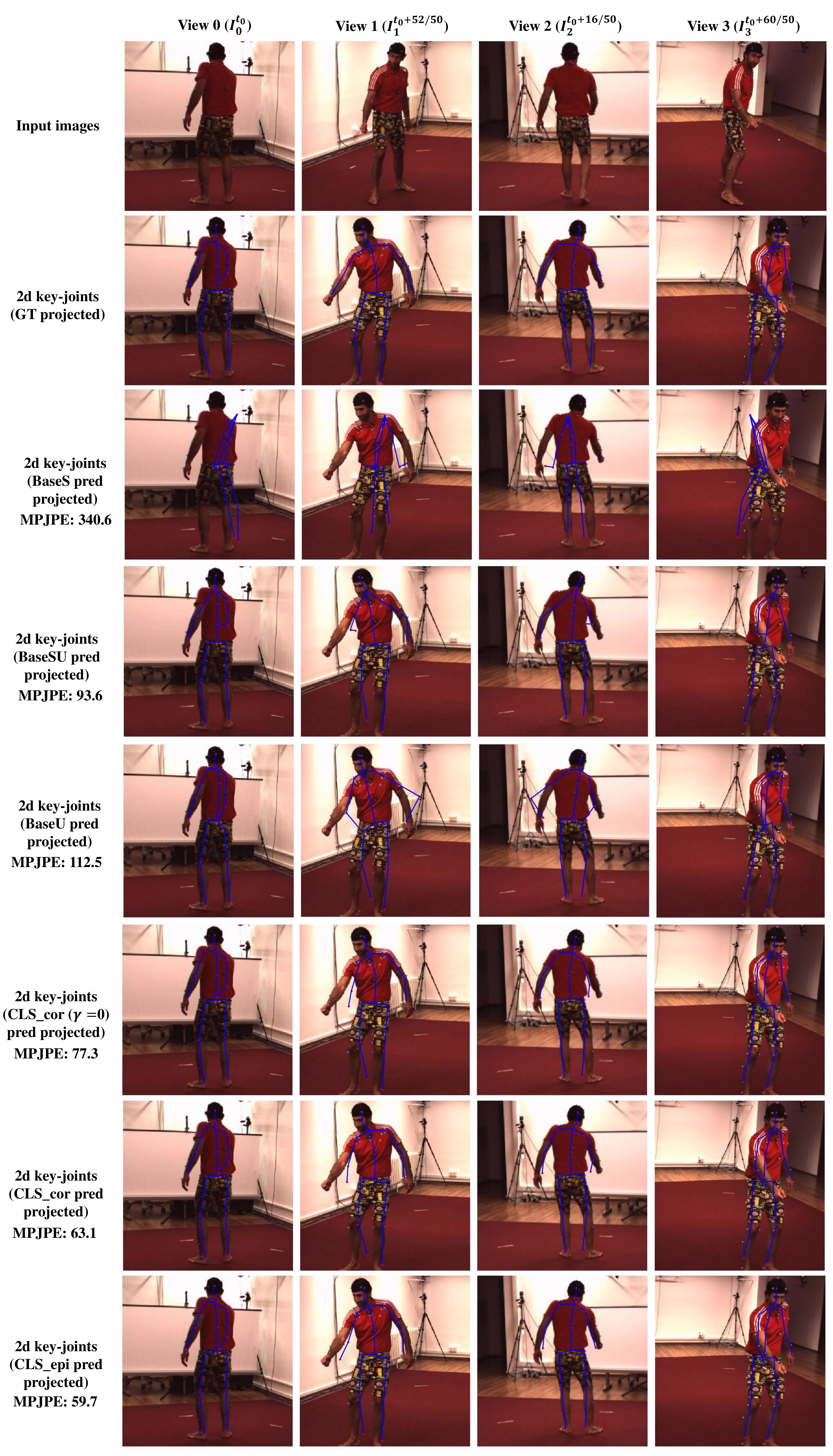}
\caption{\abc{Examples of unsynchronized 3D pose estimation (Walking Dogs). The first row shows the input unsynchronized multi-view frames and the top labels indicate the unsynchronized frame latency (in seconds). The remaining rows show the ground-truth key-joints and the predicted results.
Blue lines are the 2D key-joints projected from 3D poses, and the {\em synchronized} frames are used for better visualization. CLS-epi achieves the best performance among all methods, especially the prediction result of arms in view 0.
}
}
  \label{fig:3d_pose_estimation_example1}
\end{figure*}

\begin{figure*}[t]
  \centering
\includegraphics[width=0.7\linewidth]{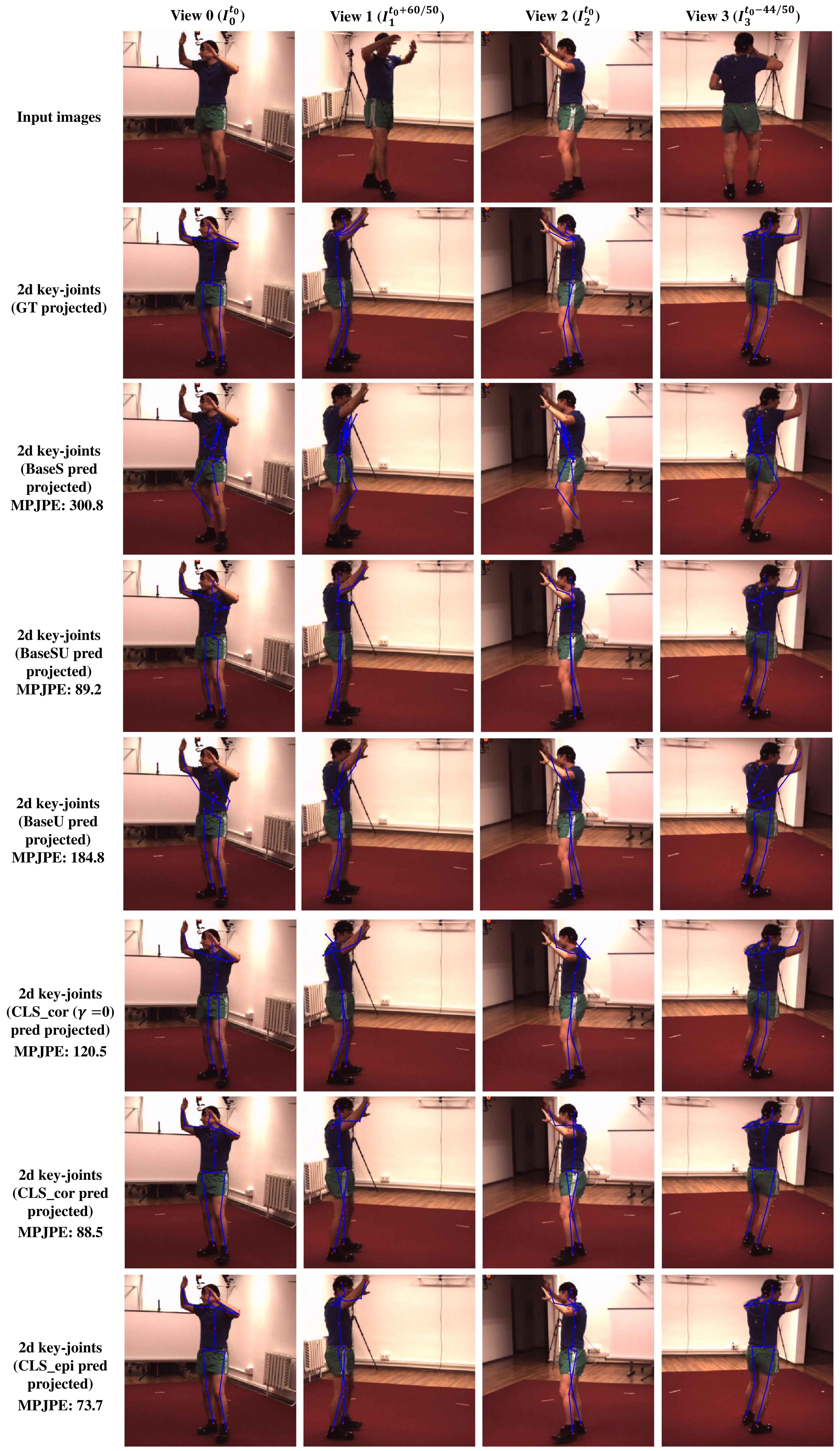}
\caption{\abc{Examples of unsynchronized 3D pose estimation (Greeting). Blue lines are the 2D
key-joints projected from 3D poses, and the synchronized frames are used for better visualization.
CLS-epi achieves the best performance.
}}
  \label{fig:3d_pose_estimation_example2}
\end{figure*}

\subsubsection{Experiment results}
 The experiments results 
are presented in Table \ref{table:results_3d_pose}. 
\Qi{
The original 3D pose estimation method (BaseS, \zq{BaseSU} and BaseU) cannot perform well under the unsynchronized test condition, \abcnn{especially under large latencies (\emph{e.g.}, 64/50s).}
 Our camera-view synchronization methods performs better than the baseline methods, \abcnn{with the performance gap increasing as the latency increases.}
Using similarity loss $\ell_s$ improves the performance of our models, and adding epipolar-guided weights can suppress false matches and further reduces the error.}
The detailed performance for each pose type under different frame latency settings is shown in Table \ref{table:results_3d_pose_8}, Table \ref{table:results_3d_pose_32} and Table \ref{table:results_3d_pose_64}. From the tables, we can find that the proposed methods can perform especially better on the poses with larger movement between unsynchronized frames, \abcn{\emph{e.g.}, Walking, WalkingDogs and WalkingTogether.}

\subsubsection{Ablation study on $\gamma$ for 3D pose estimation}
The ablation study on hyperparameter $\gamma$ for the method CLS-epi for 3D pose estimation is presented in Table \ref{table:gamma}. In general, $\gamma=0.01$ achieves better performance than other weights.

\subsubsection{Model size and running speed comparison}
\zq{
We present the model sizes and running speed comparisons of our proposed models and the baselines
for 3D pose estimation in Table \ref{tab:speed2}.
The input resolution for the correlation step of the camera-view synchronization module is $48\times48$.
As the original synchronized 3D pose estimation model \cite{iskakov2019learnable} is already very large, the running speed of the proposed models CLS-cor and CLS-epi is similar to the baseline methods BaseS/BaseSU/BaseU.
}


\begin{table}
\centering
\caption{The model parameter number and running speed comparison of the baseline methods BaseS/BaseSU/BaseU and the proposed CLS-cor and CLS-epi for 3D pose estimation on the Human3.6M dataset. The input resolution for the correlation step of the camera-view synchronization module is $48\times48$.}
\begin{tabular}{l|c|c}
    \hline
    Method   & Paras. Num    & FPS \\
   \hline
   BaseS/BaseSU/BaseU    & 80.6M        & 3.7     \\
   \hline
   CLS-cor   & 86.3M        & 3.4 \\
   CLS-epi   & 86.3M        & 3.0 \\
   \hline

\end{tabular}

\label{tab:speed2}
\vspace{-0.2cm}

\end{table}

\subsubsection{Visualization results} 

Visualization results of unsynchronized 3D pose estimation are presented in Figs.~\ref{fig:3d_pose_estimation_example1} and \ref{fig:3d_pose_estimation_example2}.
In the figures, the first row shows the input unsynchronized multi-view frames, and the top labels indicate the unsynchronized frame latency. Rows 2-8 show the 2D key-joints projected from 3D poses of Ground-truth, BaseS, \zq{BaseSU}, BaseU, CLS-cor ($\gamma = 0$), CLS-cor and CLS-epi, respectively, where {\em synchronized} frames are displayed for better visualization effect.
In Fig. \ref{fig:3d_pose_estimation_example1}, BaseU fails on the unsynchronized input, and CLS-epi achieves the best performance among all methods, especially the prediction of the arms in view 1.
In Fig. \ref{fig:3d_pose_estimation_example2}, the CLS-epi also achieves the best performance among all comparison methods.




\section{Conclusion}
\label{text:conc}

In this paper, we focus on the issue of unsynchronized cameras in DNNs-based multi-view computer vision tasks.
We propose two view synchronization models based on single frames (not videos) from each view, scene-level synchronization and camera-level synchronization. The two models are trained and evaluated under two training settings (with or without synchronized frame pairs), and a similarity loss of the projected multi-view features is proposed to boost the performance when synchronized training pairs are not available. Furthermore, to show its generality to different conditions of desynchronization, the proposed models are tested with desynchronization based on both constant and random latency. Finally, the proposed models are applied to unsynchronized multi-view counting and unsynchronized 3D human pose estimation, and achieve better performance compared to the baseline methods.
\zq{ Overall, camera-level synchronization model using correlation and epipolar weights (CLS-epi) performs best among the proposed models.}

\par
\zq{
In addition to unsynchronized multi-camera crowd counting and 3D pose estimation, the proposed method can also be applied to other multi-camera vision tasks, such as multi-camera detection \cite{chavdarova2018wildtrack}, multi-camera tracking \cite{2020City}.
In these tasks, multi-cameras may also be unsynchronized due to no synchronization clock or limited network bandwidth.
As these DNN models \cite{chavdarova2018wildtrack, 2020City} generally follow the 3 stage pipeline (single-view feature extraction, multi-view projection and fusion, and prediction), our proposed synchronization modules can be inserted to adapt them to unsynchronized frames.
}
\par
In our current model, image content matching is used for view synchronization, while the 2D-to-3D projection for multi-view fusion relies on known camera parameters.
\zq{The multi-camera surveillance tasks themselves require known calibration for better multi-view fusion. Note that our proposed view synchronization module based on correlation maps (CLS-cor) does not require camera calibrations  due to the single-frame basis, and still achieves good performance.
When the calibrations are provided, epipolar constraint can be utilized to achieve better results (CLS-epi).
In future work, the 2D-3D projection in the original multiview models could be replaced with camera self-calibration modules, which can allow the model to handle unsynchronized and uncalibrated multi-cameras.}

\section{Acknowledgements}
This work was supported by grants from the Research Grants Council of the Hong Kong Special Administrative Region, China (Project No. [T32-101/15-R] and CityU 11212518).

\ifCLASSOPTIONcaptionsoff
  \newpage
\fi




\bibliographystyle{IEEEtran}
\bibliography{egbib}

\end{document}